\documentclass[letterpaper]{article} % DO NOT CHANGE THIS
\usepackage{aaai2027}
\usepackage{cite}
\usepackage{amsmath,amssymb,amsfonts}
\usepackage{algorithmic}
\usepackage{graphicx}
\usepackage{textcomp}
\usepackage{xcolor}

\usepackage{algorithm}
\usepackage{array}

\usepackage{booktabs}
\usepackage{tabularx}
\newcolumntype{L}[1]{>{\raggedright\arraybackslash}m{#1}}
\newcolumntype{C}[1]{>{\centering\arraybackslash}m{#1}}

\usepackage{textcomp}
\usepackage{url}
\usepackage{verbatim}

\usepackage{comment}
\usepackage{makecell}  % 支持单元格内容居中控制
\usepackage{pifont}
\usepackage[utf8]{inputenc} % allow utf-8 input
\usepackage[T1]{fontenc}    % use 8-bit T1 fonts
\usepackage{booktabs}       % professional-quality tables
\usepackage{nicefrac}       % compact symbols for 1/2, etc.
\usepackage{microtype}      % microtypography
\usepackage{xcolor}  % colors
\usepackage{enumitem}
\usepackage{subcaption}
\usepackage{multirow}
\usepackage{colortbl,xcolor}
\usepackage[table,xcdraw]{xcolor} % 支持表格着色
\usepackage{booktabs, tabularx}
\usepackage{CJKutf8}
\usepackage{mathtools}
\usepackage{amsthm}

\definecolor{yellow}{RGB}{255, 215, 0}
\definecolor{orange}{RGB}{204, 130, 51}
\definecolor{grey}{RGB}{192, 192, 192}
\definecolor{mygreen}{rgb}{0.0, 0.5, 0.0}

\usepackage{graphicx} % DO NOT CHANGE THIS
\usepackage{natbib}  % DO NOT CHANGE THIS AND DO NOT ADD ANY OPTIONS TO IT
\usepackage{caption} % DO NOT CHANGE THIS AND DO NOT ADD ANY OPTIONS TO IT
\usepackage{algorithm}
\usepackage{algorithmic}

\usepackage{newfloat}
\usepackage{listings}
\DeclareCaptionStyle{ruled}{labelfont=normalfont,labelsep=colon,strut=off} % DO NOT CHANGE THIS
\floatstyle{ruled}
\newfloat{listing}{tb}{lst}{}
\floatname{listing}{Listing}

\usepackage{booktabs}

\title{A Large-Scale Chinese Knowledge Graph–Text Alignment Dataset for Benchmarking Knowledge-Grounded LLMs}
\author{
    Chengwei Wu\textsuperscript{\rm 2},
    Xingrui Zhuo\textsuperscript{\rm 3},
    Mingyang Gao\textsuperscript{\rm 4},
    Xinghe Cheng\textsuperscript{\rm 5},
    Zhichao Yan\textsuperscript{\rm 6},
    Jiapu Wang\textsuperscript{\rm 1}
}
\affiliations{
    \textsuperscript{\rm 1}Nanjing University of Science and Technology\quad
    \textsuperscript{\rm 2}Beihang University\quad
    \textsuperscript{\rm 3}Hefei University of Technology\quad\\
    \textsuperscript{\rm 4}University of Science and Technology Beijing\quad
    \textsuperscript{\rm 5}Jinan University\quad
    \textsuperscript{\rm 6}Shanxi University\quad\\
}

\begin{document}

\maketitle

\begin{abstract}
Reliable evaluation of knowledge-grounded Large Language Models (LLMs) in Chinese requires resources that explicitly align Chinese-language text with verifiable Knowledge Graph (KG) facts. Yet existing Chinese benchmarks primarily assess general language understanding and offer limited support for structured reasoning under Chinese-specific linguistic phenomena. We introduce the \textbf{C}hinese \textbf{D}ata-\textbf{T}ext \textbf{P}air (CDTP), a large-scale Chinese KG–text alignment dataset comprising more than $7$ million aligned instances across four broad domains. Each instance pairs a Chinese-language text with one or more textually supported KG triples, totaling $15$ million triples. A multi-stage construction pipeline combining alignment filtering, manual verification, and external evidence validation improves semantic consistency and factual reliability. CDTP supports Knowledge Graph Completion (KGC), Question Answering (QA), and Triple-to-Text Generation (T2T). Across all three tasks, the benchmark design accounts for Chinese-specific phenomena, including polysemy, word-segmentation ambiguity, and context-dependent entity interpretation, enabling the evaluation of structured reasoning, ambiguity-aware factual understanding, and knowledge-grounded generation. Experiments with diverse open-source and proprietary LLMs show that model scale alone does not guarantee reliable performance on these Chinese knowledge-intensive tasks, whereas supervised fine-tuning on CDTP consistently improves in-domain performance and out-of-distribution robustness. The publicly accessible dataset, code, and evaluation protocols provide a reusable resource for developing and evaluating knowledge-grounded LLMs in Chinese.

%Despite the strong Chinese language capabilities of Large Language Models (LLMs), their ability to align textual semantics with verifiable Knowledge Graph (KG) facts remains insufficiently evaluated. Existing benchmarks mainly focus on general language understanding, offering limited support for assessing structured knowledge reasoning under Chinese-specific challenges such as polysemy, segmentation ambiguity, and context-dependent entity interpretation. To fill this gap, we introduce a large-scale Chinese knowledge graph--text alignment benchmark for knowledge-grounded LLM evaluation. Built upon the Chinese Data-Text Pair (CDTP), a corpus of aligned Chinese text and knowledge graph triples, the benchmark contains over 7 million text--triple pairs and 15 million triples across four major domains. It supports three representative tasks, namely Knowledge Graph Completion, Question Answering, and Triple-to-Text Generation, to evaluate structured reasoning, ambiguity-aware factual understanding, and knowledge-grounded generation. Extensive experiments on diverse open-source and proprietary LLMs reveal that model scale alone is insufficient for robust structured knowledge understanding, while CDTP-based supervised fine-tuning consistently improves in-domain performance and out-of-distribution robustness. We release the dataset, code, and evaluation protocol as reusable infrastructure for Chinese knowledge-grounded LLM evaluation.
\end{abstract}
%\footnote{\href{https://github.com/jiapuwang/DTP-A-Large-Scale-Chinese-Data-Text-Pair-Dataset-for-Comprehensive-Evaluation-of-LLMs.git}{Code is available: https://github.com/CDTP.git}}


\section{Introduction}\label{Intro}
Large Language Models (LLMs) \cite{pan2024unifying, chang2024survey} have demonstrated exceptional performance across a broad range of Natural Language Processing (NLP) tasks \cite{wang2024large, yan2025prompting, yan2024atomic}, positioning themselves as pivotal milestones on the path toward Artificial General Intelligence (AGI) \cite{pei2019towards, fei2022towards}. 
However, achieving AGI requires mastering not only proficiency in widely studied languages like English, but also mastery of complex linguistic systems such as Chinese, which present unique and substantial challenges.

% \begin{figure}[t]
%     \centering
%     \includegraphics[scale=0.05]{CDTP_DATA.png}
%     \caption{An illustrative example of the proposed Chinese Data-Text Pair (CDTP) dataset. 
%     Each instance aligns structured triples with a corresponding Chinese text, enabling explicit grounding between symbolic knowledge and natural language descriptions.}
%     \label{fig:data}
% \end{figure}

\begin{table}[t]
\tiny
\centering
\setlength{\tabcolsep}{2pt}
\renewcommand{\arraystretch}{1.15}

\begin{tabularx}{\columnwidth}{
@{}
>{\centering\arraybackslash}p{0.18\columnwidth}
>{\centering\arraybackslash}X
>{\centering\arraybackslash}X
@{}
}
\toprule
\textbf{Chinese Expression}
& \textbf{Interpretation 1}
& \textbf{Interpretation 2} \\
\midrule

\multirow{2}{*}{\begin{CJK}{UTF8}{gbsn}行\end{CJK}}
& \begin{CJK}{UTF8}{gbsn}行动\end{CJK}
& \begin{CJK}{UTF8}{gbsn}行业\end{CJK} \\
\cmidrule(lr){2-3}

& \texttt{to do/go}
& \texttt{profession/industry} \\
\midrule

\multirow{3}{*}{\begin{CJK}{UTF8}{gbsn}武汉市长江大桥\end{CJK}}
& \begin{CJK}{UTF8}{gbsn}武汉市长/江大桥\end{CJK}
& \begin{CJK}{UTF8}{gbsn}武汉市/长江大桥\end{CJK} \\
\cmidrule(lr){2-3}

&
\shortstack{\texttt{the mayor of Wuhan}\\
            \texttt{named Jiang Daqiao}}
&
\shortstack{\texttt{the Yangtze River Bridge}\\
            \texttt{in Wuhan}} \\
\bottomrule
\end{tabularx}
\caption{\label{tab:poly}
Examples illustrating polysemy and structural ambiguity in Chinese text.
Each case shows one surface form with two possible interpretations.
}
\vspace{-10pt}
\end{table}

% \begin{table*}[t]
% \tiny
% \centering
% \caption{\label{tab:poly} Examples illustrating polysemy and structural ambiguity in Chinese text. 
% Each case shows one surface form with two possible interpretations. 
% }
% \begin{tabular}{@{}cccc@{}}
% \toprule
% \textbf{Chinese Expression} & \textbf{Interpretation 1} &{} & \textbf{Interpretation 2}\\
% \midrule
% \multirow{2}{*}{\begin{CJK}{UTF8}{gbsn}行\end{CJK}} & {\begin{CJK}{UTF8}{gbsn}行动\end{CJK}}&{} & {\begin{CJK}{UTF8}{gbsn}行业\end{CJK}} \\
% \cline{2-4}
% {}&{\texttt{to do/go}} &{} &{\texttt{profession/industry}}\\
% \hline
% \multirow{2}{*}{{\begin{CJK}{UTF8}{gbsn}武汉市长江大桥\end{CJK}}} & {\begin{CJK}{UTF8}{gbsn}武汉市长/江大桥\end{CJK}} &{} & {\begin{CJK}{UTF8}{gbsn}武汉市/长江大桥\end{CJK}} \\
% \cline{2-4}
% {}&{\texttt{the mayor of Wuhan named Jiang Daqiao}} &{} &{\texttt{the Yangtze River Bridge in Wuhan}}\\
% \bottomrule
% \end{tabular} % Adjust spacing
% \end{table*}

Chinese poses distinctive challenges for LLMs, largely because its corpora are dominated by unstructured free text and lack explicit structured annotations such as entity–attribute–value triples, which are essential for learning reliable entity–relation mappings \cite{liu2022chinese, zhu2022improving}. Table~\ref{tab:poly} illustrates this issue: a single Chinese expression can yield multiple plausible interpretations depending on context. This absence of structured signals constrains performance on a wide range of knowledge-intensive tasks, including knowledge retrieval~\cite{li2025matching}, knowledge question-answering~\cite{DBLP:conf/aaai/HuangBHZW26}, relation inference~\cite{wang2024ime,zhuo2026knowledge}, and text generation~\cite{li2024pre}. In these settings, models are required to extract or memorize facts from lengthy unstructured text~\cite{Jing2026knowledge}, a process that increases training difficulty, propagates errors, weakens generalization across domains, and ultimately reduces interpretability.

To bridge this gap, we construct the Chinese Data-Text Pair (CDTP) dataset, which tightly aligns structured triples with corresponding natural language text. CDTP is developed through rigorous collection and cleaning procedures to guarantee strict alignment, high-quality content, and broad domain coverage. In total, it comprises over $7$ million text samples and their corresponding $15$ million triples, spanning four major domains: \textit{History and Politics}, \textit{Humanities and Society}, \textit{Technology and Economics}, and \textit{Nature and Environment}. This large-scale, systematically curated resource provides a solid foundation for evaluating the interplay between structured knowledge and unstructured text. 

\begin{figure*}[t]
    \centering
    \includegraphics[width=0.9\linewidth]{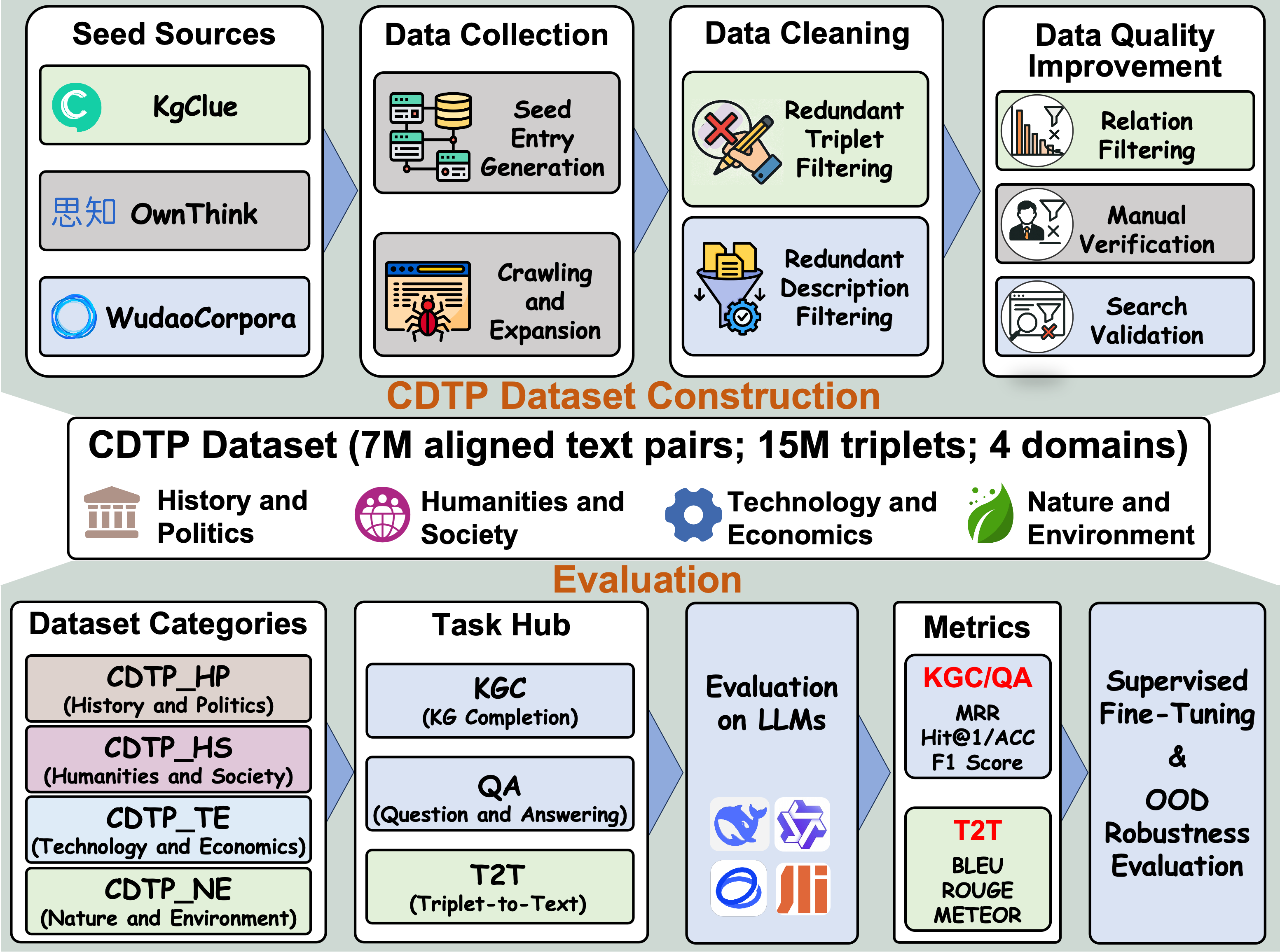}
    \caption{\label{fig:framework}Unified framework of CDTP construction and LLM evaluation. The upper part illustrates the CDTP construction pipeline, including seed source aggregation, data collection, data cleaning, and data quality improvement. The resulting dataset contains large-scale aligned text--triple pairs across four domains. Based on CDTP, the lower part presents the benchmark pipeline, including domain-specific subsets, three benchmark tasks, task-specific evaluation metrics, and downstream supervised fine-tuning and OOD robustness assessment.
    }
    \vspace{-10pt}
\end{figure*}

Moreover, existing benchmarks for evaluating LLMs, such as CMMLU \cite{li2023cmmlu}, C-Eval \cite{huang2024c}, and SuperCLUE \cite{xu2023superclue}, are largely shaped by English-centric task design and fail to capture challenges unique to Chinese, including the scarcity of structured annotations and the prevalence of ambiguous expressions. Consequently, they provide only an incomplete assessment of LLMs in knowledge-intensive tasks requiring relational reasoning and factual accuracy. To overcome these limitations, we further present the Comprehensive Benchmark for Evaluating Large Language Models, constructed on top of CDTP to provide systematic and knowledge-driven evaluation. The benchmark targets three representative tasks, including \textit{Knowledge Graph Completion (KGC)}~\cite{wang2024large, wang2023survey}, \textit{Triple-to-Text Generation (T2T)}~\cite{li2021triple}, and \textit{Question Answer (QA)}~\cite{he2024g}. Together, these tasks capture the interplay between structured knowledge and natural language, providing a more targeted evaluation of LLMs.

Building on these tasks, we conducted experiments to evaluate LLMs across three key dimensions: effectiveness, Supervised Fine-Tuning (SFT) \cite{dong-etal-2024-abilities, instag2025}, and robustness. For \textbf{effectiveness}, we evaluate the performance of $8$ LLMs across $3$ tasks on $4$ distinct sub-datasets. This evaluation assesses how well the LLMs generate accurate responses by integrating structured knowledge, and how they perform across different domains and task-specific challenges. Regarding \textbf{Supervised Fine-Tuning}, we conduct further experiments to evaluate the impact of incorporating additional labeled data on the performance of LLMs. In terms of \textbf{robustness}, we assess the stability of the SFT LLMs by testing their performance on Out-Of-Distribution (OOD) data. This aims to explore how well LLMs can maintain accuracy and reliability when faced with data that deviates from the training distribution. To summarize, the primary contributions of this paper are three-fold:
\begin{itemize}[left=0em]
\item \textbf{Comprehensive Benchmark.} We introduce a comprehensive benchmark tailored to the unique linguistic and cultural complexities of Chinese, such as context-dependent disambiguation, morphological segmentation, and idiomatic expression comprehension, addressing a critical gap in existing evaluations;
\item \textbf{High-Quality Aligned Dataset.} To the best of our knowledge, we have constructed the largest Chinese Data-to-Text Pair (CDTP) dataset, comprising 7 million meticulously aligned text pairs, each consisting of unstructured text paired with one or more corresponding triples, totaling 15 million triples across four major domains;
\item \textbf{Multi-Dimensional Evaluation.} We systematically evaluate LLMs from multiple perspectives, including performance on diverse tasks and datasets, generalizability across domains and tasks, and robustness against Out-Of-Distribution (OOD) data.
\end{itemize}

\section{Construction of the CDTP Dataset}

As illustrated in Figure~\ref{fig:framework}, the Chinese Data-to-Text Pair (CDTP) dataset is constructed through a two-stage pipeline: data collection and data processing. During data collection, we gather a raw corpus of approximately 27 million entries from large-scale online encyclopedic sources, such as \textit{Baidu Baike} and \textit{Sogou Baike}. During data processing, we perform data cleaning and quality enhancement to improve alignment consistency, factual reliability, and domain coverage. We further introduce targeted strategies to increase data diversity and support balanced evaluation across multiple domains. In the following, we describe the CDTP construction pipeline, together with its statistical characteristics and category-level organization.

\subsection{Data Collection} \label{DataCollection}

The data collection pipeline constructs a large-scale KG--text corpus from heterogeneous sources while ensuring broad coverage and high-quality alignment. The raw corpus is drawn from publicly accessible online encyclopedic sources, such as \textit{Baidu Baike} and \textit{Sogou Baike}. CDTP is developed for academic benchmark construction and evaluation, with release centered on processed benchmark instances, annotations, statistics, and code rather than large-scale raw webpage content. Source attribution is retained where applicable, and only the information necessary for dataset construction and evaluation is preserved. The pipeline comprises two stages, seed entry generation and data crawling with recursive expansion, and yields approximately 30 million triples involving around 15 million unique entities.

\textbf{Seed Entry Generation.}
We begin by aggregating seed entries from three complementary resources, namely OwnThink, KgClue, and WudaoCorpora. This step yields approximately 56 million seed entries, which serve as the basis for subsequent crawling. By combining resources with different characteristics, the seed set provides broad initial coverage across domains and entity types.

\textbf{Data Crawling and Expansion.}
Starting from the generated seed entries, we develop a customized Baike crawler to collect textual descriptions and structured relation tables from encyclopedic pages. Each page usually contains a concise entity description and an infobox-style attribute--value table, which together provide the raw materials for constructing aligned text--triple pairs. To improve coverage, we expand the crawl space by following entity hyperlinks from each seed page in a depth-first manner and collecting the textual and structured information of newly visited entities. Through this process, the initial seed set is transformed into a large-scale KG--text corpus covering a broad range of entities, relations, and semantic domains.

\subsection{Data Processing}
The data processing pipeline refines raw collections to ensure accuracy, consistency, and relevance, comprising two main stages: data cleaning and quality enhancement. In the cleaning stage, we remove noise and inconsistencies using string matching, position-based filtering, and rule-based deduplication, while the enhancement stage improves reliability through statistical filtering, expert validation, and external knowledge verification, ensuring precise alignment between structured triples and textual descriptions with high relational accuracy.

\textbf{Data Cleaning.} Given the encyclopedic nature of Baike-sourced text data, we specifically target two prevalent types of noise in our cleaning process: (1) \textit{redundant triples}—relation assertions unsupported by textual evidence, and (2) \textit{redundant descriptions}—repetitive or irrelevant content that deviates from the corresponding structured knowledge, thus ensuring better alignment between text and triples. 

\begin{table}[h]
\centering
\tiny
\renewcommand{\arraystretch}{1.2}
\resizebox{0.48\textwidth}{!}{
\begin{tabular}{@{}cl@{}}
\toprule
& {\color[HTML]{FE0000}(\begin{CJK}{UTF8}{gbsn}《90电影金曲精选》，专辑语言，粤语\end{CJK})} \\
\multirow{-2}{*}{\textbf{Triple 1}} & {\color[HTML]{FE0000}(``Golden Songs in 90s Movies'', album\_language, Cantonese)} \\
\hline
& (\begin{CJK}{UTF8}{gbsn}《90电影金曲精选》，专辑歌手，许冠杰\end{CJK}) \\
\multirow{-2}{*}{\textbf{Triple 2}} & (``Golden Songs in 90s Movies'', album\_singer, \texttt{Guanjie Xu}) \\
\hline
& (\begin{CJK}{UTF8}{gbsn}《90电影金曲精选》，曲目数量，12首\end{CJK}) \\
\multirow{-2}{*}{\textbf{Triple 3}} & (``Golden Songs in 90s Movies'', number\_of\_songs, 12) \\
\midrule
& \begin{CJK}{UTF8}{gbsn}《90电影金曲精选》是许冠杰发行的音乐专辑，共收录\end{CJK} \\
& \begin{CJK}{UTF8}{gbsn}了12首歌曲。\end{CJK} \\
& ``Golden Songs in 90s Movies'' is a music album released by \\
\multirow{-4}{*}{\textbf{Text}} & \texttt{Guanjie Xu}, which contains 12 songs. \\
\bottomrule
\end{tabular}}
\caption{\label{tab:RT} An example of redundant triples. Triples not supported by the associated text are highlighted in red.}
\end{table}

\textit{Redundant Triples.} 
Redundant triples refer to triples that are not mentioned in the accompanying text description. As illustrated in Table \ref{tab:RT}, the triple ``(\begin{CJK}{UTF8}{gbsn}
90电影金曲精选, 专辑语言, 粤语
\end{CJK}) (Golden Songs in 90s Movies, album language, Cantonese)'' conveys factual information about the album but is not explicitly reflected in the corresponding textual description, and is therefore treated as a redundant triple. To ensure alignment, we examine each data pair by checking whether the semantic content of each triple is supported by the accompanying text. Triples that lack textual grounding are removed. If all triples in a pair are filtered out, the entire data pair is discarded. This filtering procedure ensures strong semantic alignment between structured knowledge and natural language, laying a reliable foundation for downstream tasks. Additionally, discarded triples may serve as auxiliary data for text generation or data augmentation.

\begin{table}[h]
\centering
\tiny
\renewcommand{\arraystretch}{1.2}
\resizebox{0.48\textwidth}{!}{
\begin{tabular}{@{}cl@{}}
\toprule
& (\begin{CJK}{UTF8}{gbsn}黑莓 9220，操作系统，Blackberry OS 7\end{CJK}) \\
\multirow{-2}{*}{\textbf{Triple 1}} & (BlackBerry 9220, operating\_system, Blackberry OS 7) \\
\hline
& (\begin{CJK}{UTF8}{gbsn}黑莓 9220，上市日期，2012年\end{CJK}) \\
\multirow{-2}{*}{\textbf{Triple 2}} & (BlackBerry 9220, launch\_date, 2012) \\
\midrule
& \begin{CJK}{UTF8}{gbsn}黑莓 9220是2012年上市的一款支持Blackberry OS 7操作系统\end{CJK} \\
& \begin{CJK}{UTF8}{gbsn}的{\color[HTML]{FE0000}2.4英寸}的{\color[HTML]{FE0000}3G手机}。\end{CJK} \\
& BlackBerry 9220 is a {\color[HTML]{FE0000}2.4-inch 3G mobile phone} that supports \\
\multirow{-4}{*}{\textbf{Text}} & BlackBerry Os 7 operating system, which was launched in 2012. \\
\bottomrule
\end{tabular}}
\caption{\label{tab:RD} An example of redundant descriptions. Text spans not grounded in the associated triples are highlighted in red.}
\vspace{-10pt}
\end{table}

\textit{Redundant Descriptions.}  
Redundant descriptions contain information present in the text but absent from the relation triples. 
As shown in Table~\ref{tab:RD}, text descriptions often contain additional factual details, such as ``2.4 inches'' or ``3G mobile phone'', that are not represented in the associated triples. To identify and eliminate such redundancy, we compute a position index by locating the character offset of the last matched triple and normalizing it by the total length of the text. Pairs with an index above 0.75, indicating that triples appear predominantly towards the end—are retained; otherwise, they are discarded. For the retained pairs, we inspect the remaining text beyond the last triple to detect unmatched relation mentions by cross-referencing with all Baike-extracted relations. Additionally, in enumerated structures separated by punctuation, we require that all items are grounded in the relation triples. If any listed item lacks a corresponding triple, the data pair is excluded to maintain strict alignment. 

\textbf{Data Quality Improvement.} 
To enhance data reliability and control long-tail noise in large-scale KGs, we employ three complementary quality-control strategies. \textit{Relation Filtering} removes low-frequency relation types to reduce noisy predicates and retain semantically meaningful relations. 
\textit{Manual Verification} improves annotation quality through independent review of each relation type by seven trained annotators, with double annotation achieving over 90\% agreement and filtering out low-quality triples. \textit{Search Validation} further strengthens factual grounding by retrieving the top 10 external web results for each candidate triple and retaining only those whose head and tail entities co-occur in external evidence.

% \begin{figure}[h]
% \centering
% \includegraphics[width=0.5\textwidth]{top20.png}
% \caption{\label{fig:top20}Top-20 triple relation frequencies in the CDTP dataset.}
% \end{figure}

\begin{figure}[h]
\centering
\includegraphics[width=0.46\textwidth]{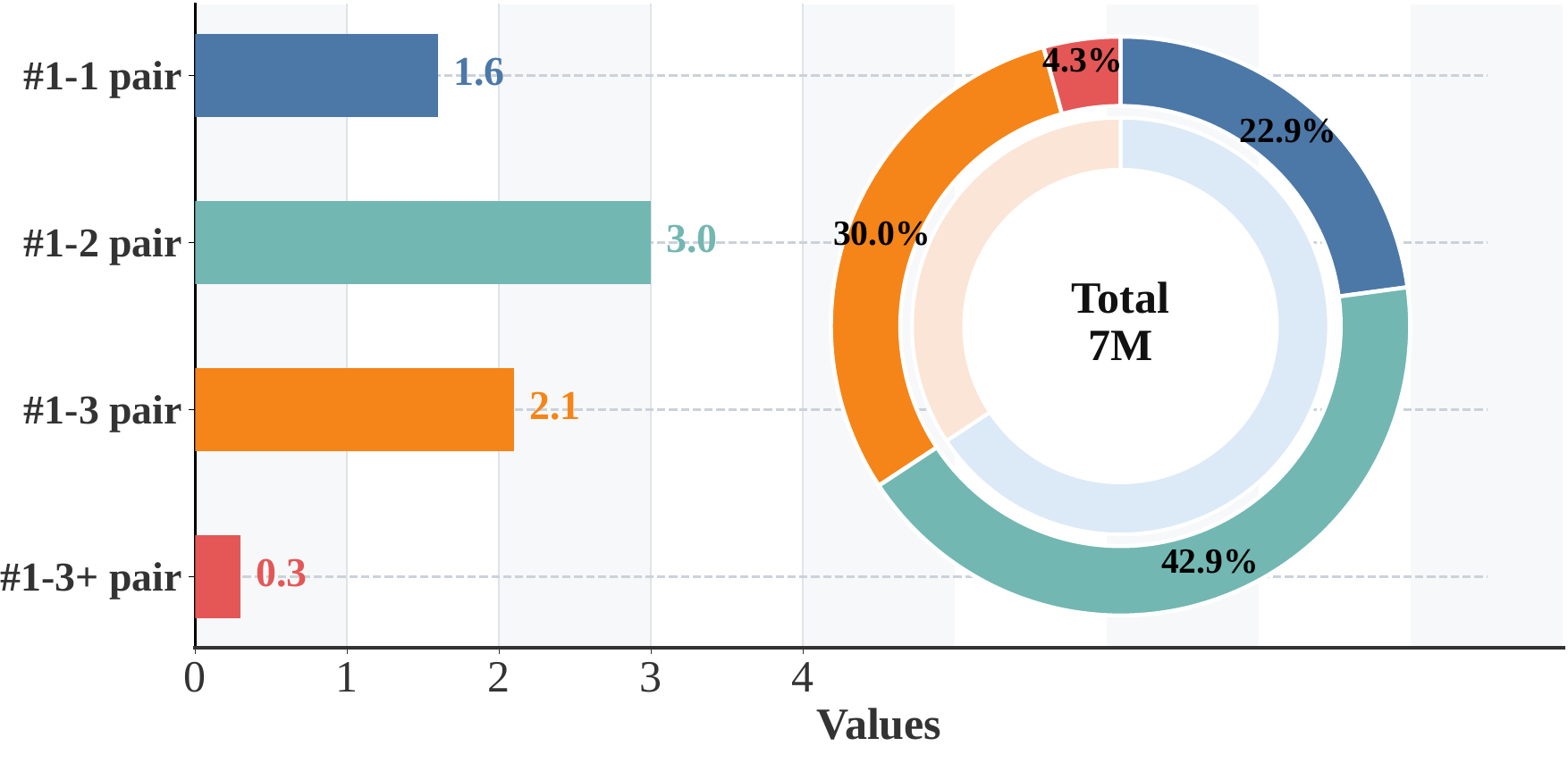}
\vspace{-5pt}
\caption{\label{fig:statistics}Distribution of data pairs by the number of associated relation triples. 
A \textbf{1-1 pair} indicates one text aligned with a single triple; 
\textbf{1-2} and \textbf{1-3} pairs follow the same convention. 
The \textbf{1-3$+$} category denotes cases where one text is aligned with more than three triples.
}
\vspace{-6pt}
\end{figure}

\textbf{Data Statistics.}
After quality control, we analyze the distribution of triple-text pairs in the curated CDTP dataset. 
As shown in Figure~\ref{fig:statistics}, CDTP contains approximately 7 million data pairs, among which \textbf{1-2 pairs} form the largest category, accounting for 42.9\% with about 3.0 million instances. 
The \textbf{1-3 pair} and \textbf{1-1 pair} categories account for 30.0\% and 22.9\%, respectively, while \textbf{1-3}$\mathbf{+}$ pairs account for only 4.3\%. 
This distribution indicates that most text descriptions are aligned with no more than three triples. 

The curated data is subsequently organized into four domains, including \textit{History and Politics}, \textit{Humanities and Society}, \textit{Technology and Economics}, and \textit{Nature and Environment}, to support domain-level evaluation. Together, these measures improve the precision, consistency, and factual validity of the dataset, providing a reliable benchmark for evaluating LLMs.

\section{Benchmark Design}
In this section, we provide a comprehensive overview, covering benchmark settings, and evaluation metrics $\&$ implementation details.

\begin{table*}[t]
\centering
\resizebox{\textwidth}{!}{
\begin{tabular}{@{}ccccccccccc@{}}
\toprule
\multicolumn{1}{c|}{Dataset} & \multicolumn{1}{c|}{Task} & \multicolumn{1}{c|}{Metrics} & GLM-4-9B & Yi-9B & Qwen1.5-7B & InternLM2-7B & Llama-3-8B & Baichuan2-7B & DeepSeek-7B & Phi-2  \\ \midrule
\rowcolor[HTML]{FBE7E6} 
\multicolumn{1}{c|}{\cellcolor[HTML]{FBE7E6}}                            & \multicolumn{1}{c|}{\cellcolor[HTML]{FBE7E6}}                         & \multicolumn{1}{c|}{\cellcolor[HTML]{FBE7E6}MRR}      & 0.4417                         & \cellcolor[HTML]{F9DF87}0.6659 & \cellcolor[HTML]{D9D9D9}0.5458 & 0.4906                         & 0.4538                         & 0.4607                         & \cellcolor[HTML]{E1CDB6}0.5364 & 0.3305 \\
\rowcolor[HTML]{FBE7E6} 
\multicolumn{1}{c|}{\cellcolor[HTML]{FBE7E6}}                            & \multicolumn{1}{c|}{\cellcolor[HTML]{FBE7E6}}                         & \multicolumn{1}{c|}{\cellcolor[HTML]{FBE7E6}Hits@1}   & 0.3445                         & \cellcolor[HTML]{F9DF87}0.5250 & \cellcolor[HTML]{D9D9D9}0.4005 & 0.3210                         & 0.2720                         & 0.2385                         & \cellcolor[HTML]{E1CDB6}0.3710 & 0.1911 \\
\rowcolor[HTML]{FBE7E6} 
\multicolumn{1}{c|}{\cellcolor[HTML]{FBE7E6}}                            & \multicolumn{1}{c|}{\multirow{-3}{*}{\cellcolor[HTML]{FBE7E6}KGC}}    & \multicolumn{1}{c|}{\cellcolor[HTML]{FBE7E6}F1 Score} & 0.5125                         & \cellcolor[HTML]{F9DF87}0.6885 & \cellcolor[HTML]{D9D9D9}0.5719 & 0.4860                         & 0.4277                         & 0.3851                         & \cellcolor[HTML]{E1CDB6}0.5412 & 0.3209 \\
\cline{2-11}
\rowcolor[HTML]{FBE7E6} 
\multicolumn{1}{c|}{\cellcolor[HTML]{FBE7E6}}                            & \multicolumn{1}{c|}{\cellcolor[HTML]{FBE7E6}}                         & \multicolumn{1}{c|}{\cellcolor[HTML]{FBE7E6}MRR}      & 0.2594                         & \cellcolor[HTML]{F9DF87}0.4818 & \cellcolor[HTML]{E1CDB6}0.3469 & \cellcolor[HTML]{D9D9D9}0.3712 & 0.2911                         & 0.3236                         & 0.2942                         & 0.3078 \\
\rowcolor[HTML]{FBE7E6} 
\multicolumn{1}{c|}{\cellcolor[HTML]{FBE7E6}}                            & \multicolumn{1}{c|}{\cellcolor[HTML]{FBE7E6}}                         & \multicolumn{1}{c|}{\cellcolor[HTML]{FBE7E6}ACC}   & 0.1870                         & \cellcolor[HTML]{F9DF87}0.3695 & \cellcolor[HTML]{E1CDB6}0.2425 & \cellcolor[HTML]{D9D9D9}0.2475 & 0.1835                         & 0.1695                         & 0.2005                         & 0.1970 \\
\rowcolor[HTML]{FBE7E6} 
\multicolumn{1}{c|}{\cellcolor[HTML]{FBE7E6}}                            & \multicolumn{1}{c|}{\multirow{-3}{*}{\cellcolor[HTML]{FBE7E6}QA}}     & \multicolumn{1}{c|}{\cellcolor[HTML]{FBE7E6}F1 Score} & 0.3151                         & \cellcolor[HTML]{F9DF87}0.5396 & \cellcolor[HTML]{E1CDB6}0.3903 & \cellcolor[HTML]{D9D9D9}0.3968 & 0.3101                         & 0.2899                         & 0.3340                         & 0.3292 \\
\cline{2-11}
\rowcolor[HTML]{FBE7E6} 
\multicolumn{1}{c|}{\cellcolor[HTML]{FBE7E6}}                            & \multicolumn{1}{c|}{\cellcolor[HTML]{FBE7E6}}                         & \multicolumn{1}{c|}{\cellcolor[HTML]{FBE7E6}BLEU}     & \cellcolor[HTML]{E1CDB6}0.2405 & \cellcolor[HTML]{F9DF87}0.2647 & 0.1471                         & 0.2211                         & 0.2058                         & 0.1542                         & \cellcolor[HTML]{D9D9D9}0.2502 & 0.1089 \\
\rowcolor[HTML]{FBE7E6} 
\multicolumn{1}{c|}{\cellcolor[HTML]{FBE7E6}}                            & \multicolumn{1}{c|}{\cellcolor[HTML]{FBE7E6}}                         & \multicolumn{1}{c|}{\cellcolor[HTML]{FBE7E6}ROUGE-1} & \cellcolor[HTML]{D9D9D9}0.7073 & \cellcolor[HTML]{F9DF87}0.7253 & 0.5516                         & \cellcolor[HTML]{E1CDB6}0.6968 & 0.6784                         & 0.5569                         & 0.6767                         & 0.5680 \\
\rowcolor[HTML]{FBE7E6} 
\multicolumn{1}{c|}{\cellcolor[HTML]{FBE7E6}}                            & \multicolumn{1}{c|}{\cellcolor[HTML]{FBE7E6}}                         & \multicolumn{1}{c|}{\cellcolor[HTML]{FBE7E6}ROUGE-L} & \cellcolor[HTML]{D9D9D9}0.6448 & \cellcolor[HTML]{F9DF87}0.6655 & 0.4969                         & 0.6223                         & 0.6099                         & 0.5100                         & \cellcolor[HTML]{E1CDB6}0.6246 & 0.5235 \\
\rowcolor[HTML]{FBE7E6} 
\multicolumn{1}{c|}{\multirow{-10}{*}{\cellcolor[HTML]{FBE7E6}CDTP\_HP}} & \multicolumn{1}{c|}{\multirow{-4}{*}{\cellcolor[HTML]{FBE7E6}T2T}} & \multicolumn{1}{c|}{\cellcolor[HTML]{FBE7E6}METEOR}   & \cellcolor[HTML]{D9D9D9}0.6972 & \cellcolor[HTML]{F9DF87}0.7030 & 0.4029                         & \cellcolor[HTML]{E1CDB6}0.6608 & 0.6462                         & 0.4872                         & 0.6442                         & 0.4556 \\ \midrule
\rowcolor[HTML]{E6E6FD} 
\multicolumn{1}{c|}{\cellcolor[HTML]{E6E6FD}}                            & \multicolumn{1}{c|}{\cellcolor[HTML]{E6E6FD}}                         & \multicolumn{1}{c|}{\cellcolor[HTML]{E6E6FD}MRR}      & 0.4927                         & \cellcolor[HTML]{F9DF87}0.6805 & \cellcolor[HTML]{E1CDB6}0.5363 & 0.4613                         & 0.4449                         & 0.4457                         & \cellcolor[HTML]{D9D9D9}0.5610 & 0.2959 \\
\rowcolor[HTML]{E6E6FD} 
\multicolumn{1}{c|}{\cellcolor[HTML]{E6E6FD}}                            & \multicolumn{1}{c|}{\cellcolor[HTML]{E6E6FD}}                         & \multicolumn{1}{c|}{\cellcolor[HTML]{E6E6FD}Hits@1}   & 0.3937                         & \cellcolor[HTML]{F9DF87}0.5430 & \cellcolor[HTML]{E1CDB6}0.3975 & 0.2925                         & 0.2640                         & 0.2070                         & \cellcolor[HTML]{D9D9D9}0.4140 & 0.1621 \\
\rowcolor[HTML]{E6E6FD} 
\multicolumn{1}{c|}{\cellcolor[HTML]{E6E6FD}}                            & \multicolumn{1}{c|}{\multirow{-3}{*}{\cellcolor[HTML]{E6E6FD}KGC}}    & \multicolumn{1}{c|}{\cellcolor[HTML]{E6E6FD}F1 Score} & 0.5650                         & \cellcolor[HTML]{F9DF87}0.7038 & \cellcolor[HTML]{E1CDB6}0.5689 & 0.4526                         & 0.4177                         & 0.3430                         & \cellcolor[HTML]{D9D9D9}0.5856 & 0.2790 \\
\cline{2-11}
\rowcolor[HTML]{E6E6FD} 
\multicolumn{1}{c|}{\cellcolor[HTML]{E6E6FD}}                            & \multicolumn{1}{c|}{\cellcolor[HTML]{E6E6FD}}                         & \multicolumn{1}{c|}{\cellcolor[HTML]{E6E6FD}MRR}      & 0.2319                         & \cellcolor[HTML]{F9DF87}0.4535 & \cellcolor[HTML]{E1CDB6}0.3072 & \cellcolor[HTML]{D9D9D9}0.3555 & 0.2915                         & 0.3049                         & 0.3012                         & 0.2932 \\
\rowcolor[HTML]{E6E6FD} 
\multicolumn{1}{c|}{\cellcolor[HTML]{E6E6FD}}                            & \multicolumn{1}{c|}{\cellcolor[HTML]{E6E6FD}}                         & \multicolumn{1}{c|}{\cellcolor[HTML]{E6E6FD}ACC}   & 0.1600                         & \cellcolor[HTML]{F9DF87}0.3310 & 0.1960                         & \cellcolor[HTML]{D9D9D9}0.2210 & 0.1675                         & 0.1352                         & \cellcolor[HTML]{E1CDB6}0.1990 & 0.1715 \\
\rowcolor[HTML]{E6E6FD} 
\multicolumn{1}{c|}{\cellcolor[HTML]{E6E6FD}}                            & \multicolumn{1}{c|}{\multirow{-3}{*}{\cellcolor[HTML]{E6E6FD}QA}}     & \multicolumn{1}{c|}{\cellcolor[HTML]{E6E6FD}F1 Score} & 0.2759                         & \cellcolor[HTML]{F9DF87}0.4974 & 0.3278                         & \cellcolor[HTML]{D9D9D9}0.3620 & 0.2869                         & 0.2382                         & \cellcolor[HTML]{E1CDB6}0.3319 & 0.2928 \\
\cline{2-11}
\rowcolor[HTML]{E6E6FD} 
\multicolumn{1}{c|}{\cellcolor[HTML]{E6E6FD}}                            & \multicolumn{1}{c|}{\cellcolor[HTML]{E6E6FD}}                         & \multicolumn{1}{c|}{\cellcolor[HTML]{E6E6FD}BLEU}     & \cellcolor[HTML]{D9D9D9}0.2235 & \cellcolor[HTML]{F9DF87}0.2556 & 0.1361                         & 0.2228                         & 0.1964                         & 0.1481                         & \cellcolor[HTML]{E1CDB6}0.2230 & 0.1012 \\
\rowcolor[HTML]{E6E6FD} 
\multicolumn{1}{c|}{\cellcolor[HTML]{E6E6FD}}                            & \multicolumn{1}{c|}{\cellcolor[HTML]{E6E6FD}}                         & \multicolumn{1}{c|}{\cellcolor[HTML]{E6E6FD}ROUGE-1} & \cellcolor[HTML]{E1CDB6}0.6992 & \cellcolor[HTML]{F9DF87}0.7174 & 0.5527                         & \cellcolor[HTML]{D9D9D9}0.7009 & 0.6694                         & 0.5533                         & 0.6628                         & 0.5706 \\
\rowcolor[HTML]{E6E6FD} 
\multicolumn{1}{c|}{\cellcolor[HTML]{E6E6FD}}                            & \multicolumn{1}{c|}{\cellcolor[HTML]{E6E6FD}}                         & \multicolumn{1}{c|}{\cellcolor[HTML]{E6E6FD}ROUGE-L} & \cellcolor[HTML]{D9D9D9}0.6408 & \cellcolor[HTML]{F9DF87}0.6586 & 0.5025                         & \cellcolor[HTML]{E1CDB6}0.6307 & 0.6079                         & 0.5054                         & 0.6128                         & 0.5181 \\
\rowcolor[HTML]{E6E6FD} 
\multicolumn{1}{c|}{\multirow{-10}{*}{\cellcolor[HTML]{E6E6FD}CDTP\_HS}} & \multicolumn{1}{c|}{\multirow{-4}{*}{\cellcolor[HTML]{E6E6FD}T2T}} & \multicolumn{1}{c|}{\cellcolor[HTML]{E6E6FD}METEOR}   & \cellcolor[HTML]{D9D9D9}0.6910 & \cellcolor[HTML]{F9DF87}0.7086 & 0.4052                         & \cellcolor[HTML]{E1CDB6}0.6696 & 0.6411                         & 0.4771                         & 0.6201                         & 0.4614 \\ \midrule
\rowcolor[HTML]{EBFEE8} 
\multicolumn{1}{c|}{\cellcolor[HTML]{EBFEE8}}                            & \multicolumn{1}{c|}{\cellcolor[HTML]{EBFEE8}}                         & \multicolumn{1}{c|}{\cellcolor[HTML]{EBFEE8}MRR}      & 0.4693                         & \cellcolor[HTML]{F9DF87}0.7249 & \cellcolor[HTML]{E1CDB6}0.6002 & 0.5105                         & 0.4527                         & 0.4602                         & \cellcolor[HTML]{D9D9D9}0.6413 & 0.2805 \\
\rowcolor[HTML]{EBFEE8} 
\multicolumn{1}{c|}{\cellcolor[HTML]{EBFEE8}}                            & \multicolumn{1}{c|}{\cellcolor[HTML]{EBFEE8}}                         & \multicolumn{1}{c|}{\cellcolor[HTML]{EBFEE8}Hits@1}   & 0.3900                         & \cellcolor[HTML]{F9DF87}0.6055 & \cellcolor[HTML]{E1CDB6}0.4735 & 0.3460                         & 0.2690                         & 0.1940                         & \cellcolor[HTML]{D9D9D9}0.5145 & 0.1490 \\
\rowcolor[HTML]{EBFEE8} 
\multicolumn{1}{c|}{\cellcolor[HTML]{EBFEE8}}                            & \multicolumn{1}{c|}{\multirow{-3}{*}{\cellcolor[HTML]{EBFEE8}KGC}}    & \multicolumn{1}{c|}{\cellcolor[HTML]{EBFEE8}F1 Score} & 0.5612                         & \cellcolor[HTML]{F9DF87}0.7543 & \cellcolor[HTML]{E1CDB6}0.6427 & 0.5141                         & 0.4240                         & 0.3249                         & \cellcolor[HTML]{D9D9D9}0.6794 & 0.2594 \\
\cline{2-11}
\rowcolor[HTML]{EBFEE8} 
\multicolumn{1}{c|}{\cellcolor[HTML]{EBFEE8}}                            & \multicolumn{1}{c|}{\cellcolor[HTML]{EBFEE8}}                         & \multicolumn{1}{c|}{\cellcolor[HTML]{EBFEE8}MRR}      & 0.3261                         & \cellcolor[HTML]{F9DF87}0.5106 & 0.3149                         & \cellcolor[HTML]{D9D9D9}0.4215 & 0.3385                         & \cellcolor[HTML]{E1CDB6}0.4023 & 0.3532                         & 0.3028 \\
\rowcolor[HTML]{EBFEE8} 
\multicolumn{1}{c|}{\cellcolor[HTML]{EBFEE8}}                            & \multicolumn{1}{c|}{\cellcolor[HTML]{EBFEE8}}                         & \multicolumn{1}{c|}{\cellcolor[HTML]{EBFEE8}ACC}   & 0.2650                         & \cellcolor[HTML]{F9DF87}0.4235 & 0.2320                         & \cellcolor[HTML]{D9D9D9}0.3145 & 0.2245                         & \cellcolor[HTML]{E1CDB6}0.2805 & 0.2750                         & 0.1925 \\
\rowcolor[HTML]{EBFEE8} 
\multicolumn{1}{c|}{\cellcolor[HTML]{EBFEE8}}                            & \multicolumn{1}{c|}{\multirow{-3}{*}{\cellcolor[HTML]{EBFEE8}QA}}     & \multicolumn{1}{c|}{\cellcolor[HTML]{EBFEE8}F1 Score} & 0.4190                         & \cellcolor[HTML]{F9DF87}0.5950 & 0.3766                         & \cellcolor[HTML]{D9D9D9}0.4785 & 0.3667                         & \cellcolor[HTML]{E1CDB6}0.4381 & 0.4314                         & 0.3229 \\
\cline{2-11}
\rowcolor[HTML]{EBFEE8} 
\multicolumn{1}{c|}{\cellcolor[HTML]{EBFEE8}}                            & \multicolumn{1}{c|}{\cellcolor[HTML]{EBFEE8}}                         & \multicolumn{1}{c|}{\cellcolor[HTML]{EBFEE8}BLEU}     & \cellcolor[HTML]{E1CDB6}0.2146 & \cellcolor[HTML]{F9DF87}0.2213 & 0.1213                         & 0.1786                         & 0.2079                         & 0.1209                         & \cellcolor[HTML]{D9D9D9}0.2153 & 0.0851 \\
\rowcolor[HTML]{EBFEE8} 
\multicolumn{1}{c|}{\cellcolor[HTML]{EBFEE8}}                            & \multicolumn{1}{c|}{\cellcolor[HTML]{EBFEE8}}                         & \multicolumn{1}{c|}{\cellcolor[HTML]{EBFEE8}ROUGE-1} & \cellcolor[HTML]{D9D9D9}0.6910 & \cellcolor[HTML]{F9DF87}0.6912 & 0.5399                         & 0.6667                         & \cellcolor[HTML]{E1CDB6}0.6789 & 0.5412                         & 0.6508                         & 0.5502 \\
\rowcolor[HTML]{EBFEE8} 
\multicolumn{1}{c|}{\cellcolor[HTML]{EBFEE8}}                            & \multicolumn{1}{c|}{\cellcolor[HTML]{EBFEE8}}                         & \multicolumn{1}{c|}{\cellcolor[HTML]{EBFEE8}ROUGE-L} & \cellcolor[HTML]{F9DF87}0.6535 & \cellcolor[HTML]{D9D9D9}0.6505 & 0.5029                         & 0.6147                         & \cellcolor[HTML]{E1CDB6}0.6367 & 0.5076                         & 0.6210                         & 0.5108 \\
\rowcolor[HTML]{EBFEE8} 
\multicolumn{1}{c|}{\multirow{-10}{*}{\cellcolor[HTML]{EBFEE8}CDTP\_NE}} & \multicolumn{1}{c|}{\multirow{-4}{*}{\cellcolor[HTML]{EBFEE8}T2T}} & \multicolumn{1}{c|}{\cellcolor[HTML]{EBFEE8}METEOR}   & \cellcolor[HTML]{F9DF87}0.6791 & \cellcolor[HTML]{D9D9D9}0.6705 & 0.4008                         & 0.6225                         & \cellcolor[HTML]{E1CDB6}0.6502 & 0.4693                         & 0.6136                         & 0.4259 \\ \midrule
\rowcolor[HTML]{FEFCE7} 
\multicolumn{1}{c|}{\cellcolor[HTML]{FEFCE7}}                            & \multicolumn{1}{c|}{\cellcolor[HTML]{FEFCE7}}                         & \multicolumn{1}{c|}{\cellcolor[HTML]{FEFCE7}MRR}      & 0.4856                         & \cellcolor[HTML]{F9DF87}0.6705 & \cellcolor[HTML]{E1CDB6}0.4973 & 0.4619                         & 0.4424                         & 0.4488                         & \cellcolor[HTML]{D9D9D9}0.5305 & 0.3112 \\
\rowcolor[HTML]{FEFCE7} 
\multicolumn{1}{c|}{\cellcolor[HTML]{FEFCE7}}                            & \multicolumn{1}{c|}{\cellcolor[HTML]{FEFCE7}}                         & \multicolumn{1}{c|}{\cellcolor[HTML]{FEFCE7}Hits@1}   & \cellcolor[HTML]{D9D9D9}0.3910 & \cellcolor[HTML]{F9DF87}0.5375 & 0.3620                         & 0.2840                         & 0.2610                         & 0.2240                         & \cellcolor[HTML]{E1CDB6}0.3675 & 0.1856 \\
\rowcolor[HTML]{FEFCE7} 
\multicolumn{1}{c|}{\cellcolor[HTML]{FEFCE7}}                            & \multicolumn{1}{c|}{\multirow{-3}{*}{\cellcolor[HTML]{FEFCE7}KGC}}    & \multicolumn{1}{c|}{\cellcolor[HTML]{FEFCE7}F1 Score} & \cellcolor[HTML]{D9D9D9}0.5622 & \cellcolor[HTML]{F9DF87}0.6992 & 0.5316                         & 0.4424                         & 0.4140                         & 0.3660                         & \cellcolor[HTML]{E1CDB6}0.5375 & 0.3130 \\
\cline{2-11}
\rowcolor[HTML]{FEFCE7} 
\multicolumn{1}{c|}{\cellcolor[HTML]{FEFCE7}}                            & \multicolumn{1}{c|}{\cellcolor[HTML]{FEFCE7}}                         & \multicolumn{1}{c|}{\cellcolor[HTML]{FEFCE7}MRR}      & 0.3070                         & \cellcolor[HTML]{F9DF87}0.5076 & 0.3486                         & \cellcolor[HTML]{D9D9D9}0.4045 & 0.3147                         & \cellcolor[HTML]{E1CDB6}0.3620 & 0.3514                         & 0.3130 \\
\rowcolor[HTML]{FEFCE7} 
\multicolumn{1}{c|}{\cellcolor[HTML]{FEFCE7}}                            & \multicolumn{1}{c|}{\cellcolor[HTML]{FEFCE7}}                         & \multicolumn{1}{c|}{\cellcolor[HTML]{FEFCE7}ACC}   & 0.2199                         & \cellcolor[HTML]{F9DF87}0.4060 & 0.2425                         & \cellcolor[HTML]{D9D9D9}0.2775 & 0.1910                         & 0.2075                         & \cellcolor[HTML]{E1CDB6}0.2585 & 0.1900 \\
\rowcolor[HTML]{FEFCE7} 
\multicolumn{1}{c|}{\cellcolor[HTML]{FEFCE7}}                            & \multicolumn{1}{c|}{\multirow{-3}{*}{\cellcolor[HTML]{FEFCE7}QA}}     & \multicolumn{1}{c|}{\cellcolor[HTML]{FEFCE7}F1 Score} & 0.3605                         & \cellcolor[HTML]{F9DF87}0.5775 & 0.3903                         & \cellcolor[HTML]{D9D9D9}0.4344 & 0.3207                         & 0.3437                         & \cellcolor[HTML]{E1CDB6}0.4108 & 0.3193 \\
\cline{2-11}
\rowcolor[HTML]{FEFCE7} 
\multicolumn{1}{c|}{\cellcolor[HTML]{FEFCE7}}                            & \multicolumn{1}{c|}{\cellcolor[HTML]{FEFCE7}}                         & \multicolumn{1}{c|}{\cellcolor[HTML]{FEFCE7}BLEU}     & \cellcolor[HTML]{D9D9D9}0.2600 & \cellcolor[HTML]{F9DF87}0.2740 & 0.1665                         & 0.2395                         & 0.2446                         & 0.1550                         & \cellcolor[HTML]{E1CDB6}0.2552 & 0.1321 \\
\rowcolor[HTML]{FEFCE7} 
\multicolumn{1}{c|}{\cellcolor[HTML]{FEFCE7}}                            & \multicolumn{1}{c|}{\cellcolor[HTML]{FEFCE7}}                         & \multicolumn{1}{c|}{\cellcolor[HTML]{FEFCE7}ROUGE-1} & \cellcolor[HTML]{D9D9D9}0.7169 & \cellcolor[HTML]{F9DF87}0.7295 & 0.5875                         & \cellcolor[HTML]{E1CDB6}0.7077 & 0.7031                         & 0.5409                         & 0.6886                         & 0.6032 \\
\rowcolor[HTML]{FEFCE7} 
\multicolumn{1}{c|}{\cellcolor[HTML]{FEFCE7}}                            & \multicolumn{1}{c|}{\cellcolor[HTML]{FEFCE7}}                         & \multicolumn{1}{c|}{\cellcolor[HTML]{FEFCE7}ROUGE-L} & \cellcolor[HTML]{D9D9D9}0.6623 & \cellcolor[HTML]{F9DF87}0.6715 & 0.5364                         & 0.6415                         & \cellcolor[HTML]{E1CDB6}0.6440 & 0.5003                         & 0.6406                         & 0.5584 \\
\rowcolor[HTML]{FEFCE7} 
\multicolumn{1}{c|}{\multirow{-10}{*}{\cellcolor[HTML]{FEFCE7}CDTP\_TE}} & \multicolumn{1}{c|}{\multirow{-4}{*}{\cellcolor[HTML]{FEFCE7}T2T}} & METEOR                                                & \cellcolor[HTML]{D9D9D9}0.7165 & \cellcolor[HTML]{F9DF87}0.7244 & 0.4608                         & 0.6795                         & \cellcolor[HTML]{E1CDB6}0.6854 & 0.4643                         & 0.6638                         & 0.4949 \\ \bottomrule
\end{tabular}}
\caption{Performance of eight LLMs on four datasets. The best three results are highlighted by \colorbox{yellow!50}{1st}, \colorbox{orange!50}{2nd}, and \colorbox{grey!50}{3rd}.}
\label{tab:eva1}
\end{table*}

\subsection{Benchmark Settings}\label{set}

\textbf{Benchmark Datasets.} To evaluate model performance across diverse domains, we construct our benchmark using four representative sub-datasets drawn from the full CDTP collection (7M pairs): \textit{CDTP\_HP (History and Politics)}, \textit{CDTP\_HS (Humanities and Society)}, \textit{CDTP\_TE (Technology and Economics)}, and \textit{CDTP\_NE (Nature and Environment)}\textcolor{red}. For the experiments in this paper, each sub-dataset is split into 80$\%$ training and 20$\%$ evaluation. To ensure tractable and reproducible experiments while preserving data diversity, we report results on a stratified 10K subset drawn from the full CDTP dataset. This subset preserves key distributions (domain, entity type, predicate frequency, and text length), is sampled with a fixed pseudorandom seed, and follows the same 80$\%$/20$\%$ split per domain. 

\textbf{Evaluation Models.}
To evaluate the performance of Chinese LLMs on our proposed CDTP dataset, we selected eight state-of-the-art models with diverse architectures and training paradigms, including both general-purpose and Chinese-focused LLMs. The selected LLMs are GLM-4-9B \citep{glm2024chatglm}, Yi-9B \citep{young2024yi}, Qwen1.5-7B \citep{bai2023qwen}, InternLM2-7B \citep{cai2024internlm2}, Llama-3-8B \citep{grattafiori2024llama}, Baichuan2-7B \citep{yang2023baichuan}, Phi-2 \citep{javaheripi2023phi}, and DeepSeek-7B \citep{bi2024deepseek}. The supplementary material provides detailed descriptions of the Chinese LLMs evaluated in this work, including those discussed in the main text, as well as additional closed-source models (GPT-4, Claude Sonnet 4.5, and Gemini 2.5 Flash) and higher-parameter open-source models (Qwen3-32B).%Details of these Chinese LLMs are provided in Appendix~\ref{ModelSources}.

\textbf{Benchmark Tasks.}
To fully unlock the potential of the CDTP dataset, we conduct a comprehensive evaluation of Chinese LLMs based on three tasks, including Question Answer (QA), Knowledge Graph Completion (KGC), and Triple to Text Generation (T2T). For QA and KGC, each query is cast as a multiple-choice problem with one correct answer and nine distractors. Distractors are drawn from KG neighbors and type-/frequency-matched entities, filtered for semantic feasibility, and enriched with polysemy-targeted candidates to capture Chinese-specific ambiguities. This design enables these tasks to assess both relational reasoning and robustness under contextual ambiguity. %{\color{red}{Detailed task definitions in supplementary}}.

%\subsection{Evaluation Metrics and Implementation Details}\label{IMP}

%\textbf{Evaluation Metrics.}
%Several commonly used metrics are employed to evaluate the performance of Chinese LLMs across the three tasks: QA, KGC, and T2T. For QA and KGC, we use Mean Reciprocal Rank (MRR), Hits@1, and F1 Score. For the T2T task, BLEU, ROUGE, and METEOR are applied. Detailed descriptions and calculation procedures for these metrics are provided in the supplementary material.
%Detailed descriptions and calculation methods for these metrics are provided in Appendix~\ref{DofMetrics}.

%\subsection{Evaluation Metrics and Implementation Details}\label{IMP}

\textbf{Evaluation Metrics.}
%Several commonly used metrics are employed to evaluate the performance of LLMs across the three tasks: QA, KGC, and T2T. For QA and KGC, we use Mean Reciprocal Rank (MRR), Hits@1, and F1 Score. For the T2T task, BLEU, ROUGE, and METEOR are applied. %Detailed descriptions and calculation methods for these metrics are provided in Appendix~\ref{DofMetrics}.
%\subsection{Details of Evaluation Metrics}\label{DofMetrics}
Several commonly used metrics are employed to evaluate the performance of LLMs across the three tasks. For QA and KGC tasks, we use ranking and classification metrics to jointly assess answer prioritization and prediction correctness. Specifically, QA is evaluated with \textit{MRR} \cite{wang2024ime}, \textit{ACC} \cite{sun2025large}, and \textit{F1 Score} \cite{sokolova2009systematic}, while KGC is evaluated with \textit{MRR}, \textit{Hits@1} \cite{wang2023multi}, and \textit{F1 Score}. For the generation task T2T, we adopt \textit{BLEU} \cite{BLEU}, \textit{ROUGE-1}, \textit{ROUGE-L} \cite{ROUGE}, and \textit{METEOR} \cite{METEOR} to measure lexical overlap, sequence-level similarity, and overall generation quality.

\textbf{Supervised Fine-Tuning (SFT) Parameters Settings.}
To validate the quality of our proposed CDTP dataset, SFT experiments were conducted on $8$ LLMs using DeepSpeed \cite{rasley2020deepspeed}. 
Each dataset category was fine-tuned independently with the following configurations: \{Training Framework: DeepSpeed with ZeRO Stage 2; Batch Size: 8; Learning Rate: $9.65 \times 10^{-6}$ with a cosine learning rate scheduler; Max Sequence Length: 2048; Mixed Precision: BF16; Number of Epochs: 3\}. All models are fine-tuned on the SFT datasets using 8 NVIDIA H100 GPUs.

\begin{figure}[t]
  \centering
    \includegraphics[width=\linewidth]{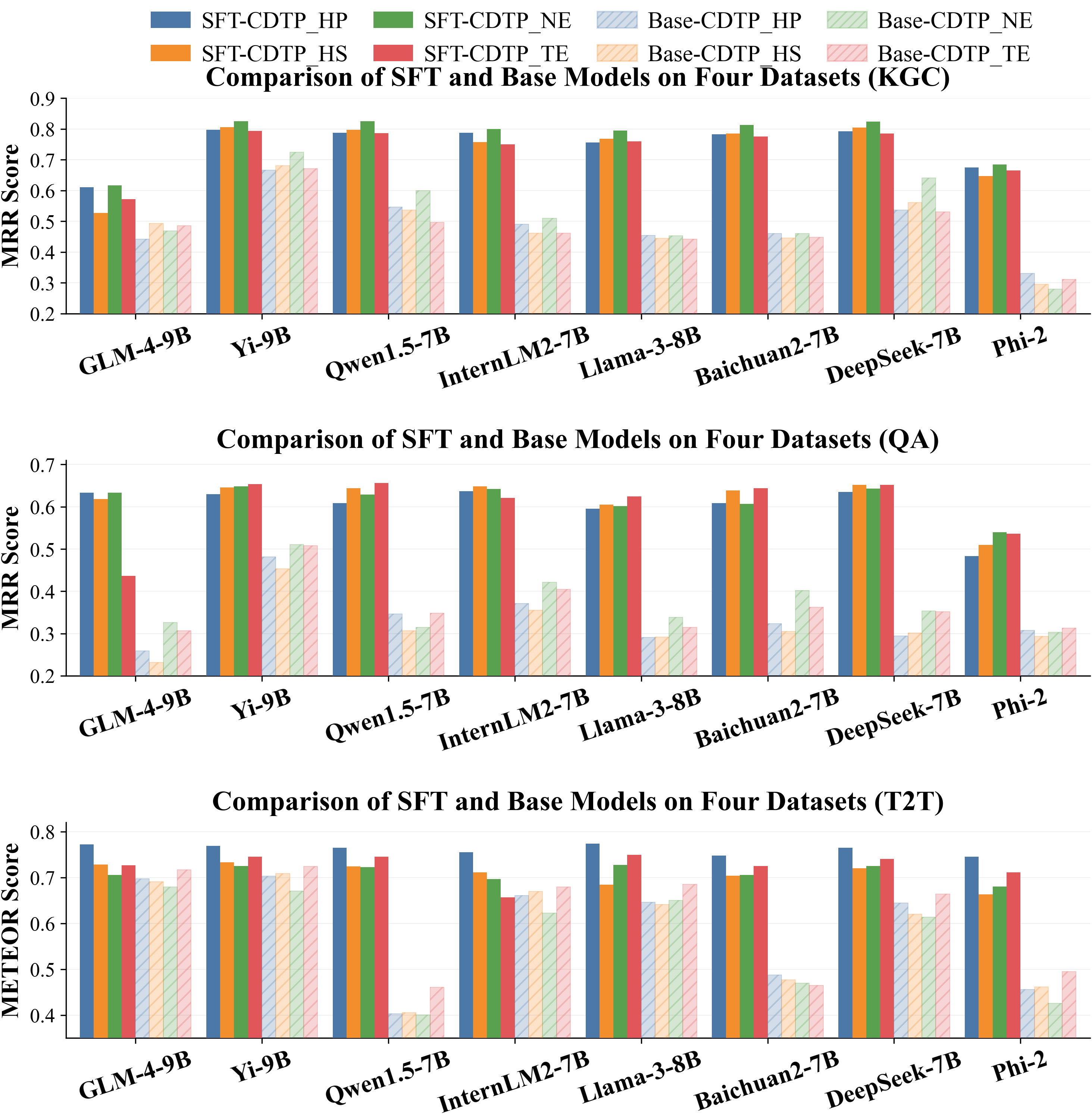}
    \vspace{-13pt}
\caption{\label{fig:Base_SFT_Stack_Chart} Performance comparison of SFT and base models on four datasets for three tasks.}
\vspace{-13pt}
\end{figure}

% \begin{figure*}[h]
%     \centering
%     \includegraphics[scale=0.29]{Ranking.png}
%     \caption{Performance ranking of eight Base and SFT models across three tasks on four datasets, namely CDTP\_HP, CDTP\_HS, CDTP\_NE, and CDTP\_TE.
%     }
%     \label{fig:Ranking}
% \end{figure*}

\section{Experimental Results}
In this section, we present the experimental results on the proposed benchmark. Specifically, we aim to answer the following research questions: \ding{108}~\textbf{RQ1 (In-domain Effectiveness):} How do different LLMs perform across tasks and datasets? \ding{108}~\textbf{RQ2 (Effect of SFT):} To what extent does supervised fine-tuning on labeled CDTP data improve the performance and cross-task generalization of LLMs?  \ding{108}~\textbf{RQ3 (Ranking and Scaling Trends):} What ranking and scaling trends can be observed among base and SFT models across tasks and datasets? \ding{108}~\textbf{RQ4 (Comparison Across Model Scales):} How do larger LLMs compare with smaller models and strong closed-source LLMs? \ding{108}~\textbf{RQ5 (OOD Robustness):} How robust are LLMs under out-of-distribution (OOD) settings, and can CDTP-based supervised fine-tuning further improve such robustness?

\subsection{RQ1: In-domain Effectiveness}\label{subsec:effectiveness}
\textbf{Experiment Design.} 
We comprehensively evaluate and compare the performance of eight LLMs across three tasks, KGC, QA, and T2T, on four distinct datasets. 

\textbf{Experimental Results.} 
Table~\ref{tab:eva1} shows the performance comparison in terms of three tasks on four datasets. We have the following observations.

\textbf{\underline{Observation \ding{182}}}: \textbf{Performance varies substantially across tasks and datasets.}
As shown in Table~\ref{tab:eva1}, strong performance on one task does not necessarily transfer to others. For example, InternLM2-7B achieves METEOR scores above 0.66 on CDTP\_TE and CDTP\_HS in T2T, yet performs less competitively on QA in terms of Hits@1 and MRR. Performance also varies across datasets within the same task; for instance, several models perform well on CDTP\_HP but poorly on CDTP\_TE in KGC, with similar patterns observed in QA and T2T. These results highlight the need for comprehensive evaluation across multiple tasks, datasets, and metrics.

\textbf{\underline{Observation \ding{183}}}: \textbf{LLMs perform better on KGC than on QA.}
As shown in Table~\ref{tab:eva1}, all evaluated LLMs achieve consistently higher MRR on KGC than on QA. This contrasts with findings in English, where LLMs often perform better on unstructured QA because of their extensive pretraining on natural-language corpora~\cite{achiam2023gpt}. The difference may reflect challenges specific to Chinese, including polysemy, syntactic ambiguity, and script variation, which can hinder unstructured language understanding. By providing explicit semantic relations, structured KGC inputs partially mitigate these ambiguities and yield more stable performance. These findings motivate evaluation methods tailored to the linguistic characteristics of Chinese.

\subsection{RQ2: Supervised Fine-Tuning (SFT) Experiments}\label{subsec:SFT}
\textbf{Experiment Design.} We perform Supervised Fine-Tuning (SFT) on these base models and subsequently reassess their performance on these datasets to measure improvements resulting from fine-tuning.

\textbf{Experimental Results.} Figure~\ref{fig:Base_SFT_Stack_Chart} shows the performance comparison in terms of three tasks on four datasets after SFT. We have the following key observations.

\textbf{\underline{Observation~\ding{184}}: Consistent improvements across tasks and datasets.}
As shown in Figure~\ref{fig:Base_SFT_Stack_Chart}, supervised fine-tuning consistently improves performance across all tasks and datasets. Yi-9B shows strong and stable transferability after fine-tuning, whereas Phi-2 and Baichuan2-7B exhibit larger gains from weaker base performance, indicating greater reliance on task-specific supervision. Overall, these results demonstrate the effectiveness of SFT in improving model generalization and task performance.

\textbf{\underline{Observation~\ding{185}}: Reduced performance gaps after fine-tuning.}
Fine-tuning narrows the performance gaps among models of different sizes and architectures, yielding more consistent results across tasks. Most fine-tuned models achieve similarly strong performance, suggesting that the training data provides broad and effective supervision. These findings further demonstrate the value of our dataset for improving model performance across diverse tasks and model types.

\subsection{RQ3: Ranking of Different Base and SFT Models}\label{Ranking}

\textbf{Experiment Design.} 
We comprehensively rank the performance of eight base and SFT LLMs across three tasks (KGC, QA, and T2T) using four distinct datasets. 

\begin{figure}[!t]
  \centering
  \begin{subfigure}[b]{\linewidth}
    \centering
    \includegraphics[width=\linewidth]{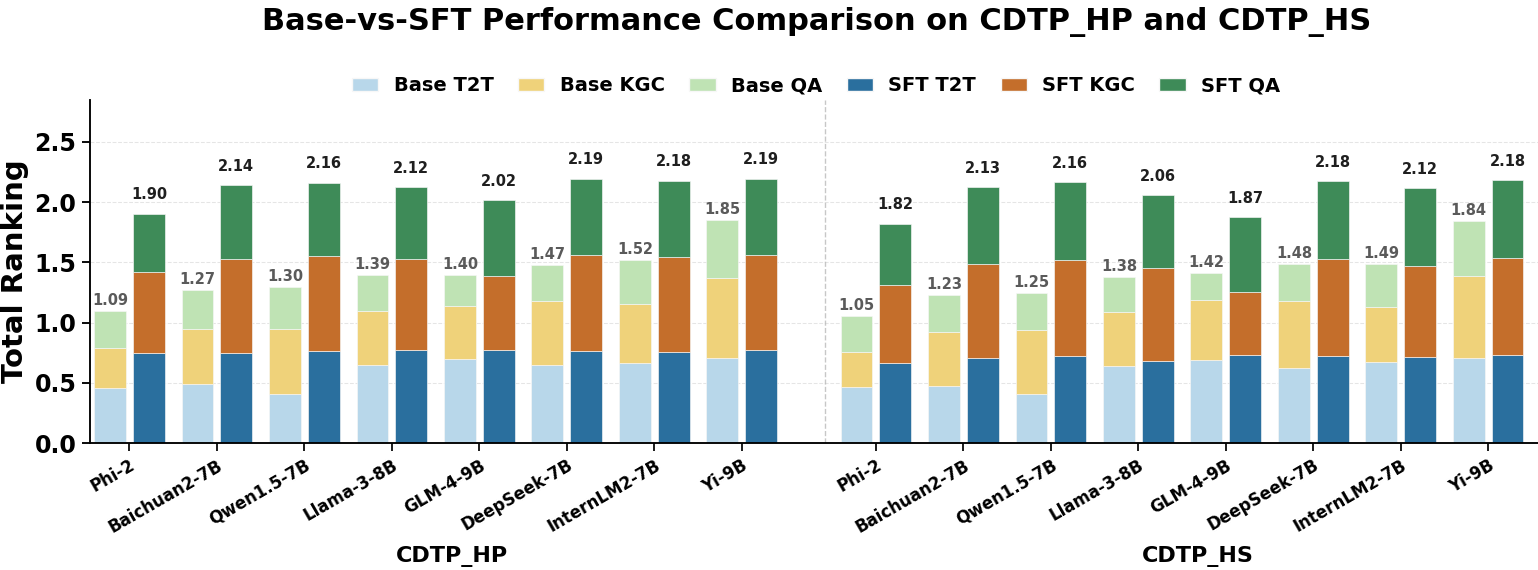}
  \end{subfigure}
  
  \begin{subfigure}[b]{\linewidth}
    \centering
    \includegraphics[width=\linewidth]{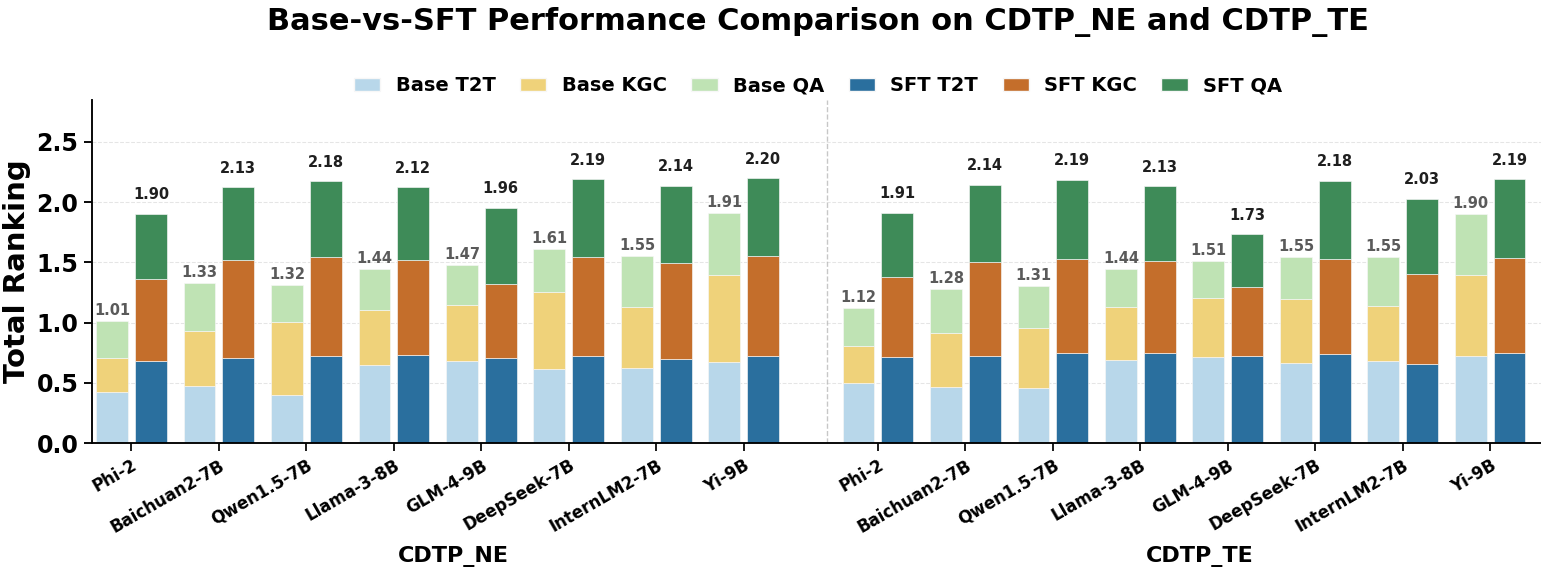}
  \end{subfigure}
  \vspace{-15pt}
\caption{Performance ranking of eight base and SFT models across three tasks on four datasets.}
\label{fig:Ranking}
\vspace{-13pt}
\end{figure}

\textbf{Experimental Results.}  Figure~\ref{fig:Ranking} provides the bar graphs to illustrate the performance rankings of LLMs with respect to their parameter sizes. We have the following observations.

\textbf{\underline{Observation \ding{186}}}: \textbf{Larger LLMs generally perform better.}
As shown in Figure~\ref{fig:Ranking}, base-model performance generally increases with model size, with Yi-9B consistently outperforming smaller models across the three tasks. This suggests that larger parameter scales improve the modeling of complex patterns and semantic relations. However, this advantage is less evident in QA, where performance depends more on reasoning and contextual understanding than on model size alone.

\textbf{\underline{Observation \ding{187}}}: \textbf{SFT substantially improves and stabilizes performance.}
Figure~\ref{fig:Ranking} shows that all fine-tuned models outperform their base counterparts. After SFT, however, the effect of model size becomes less pronounced. This indicates that SFT can partially reduce performance differences caused by parameter scale and promote more consistent performance across tasks.

% \begin{figure*}[htpb]
%   \centering
%   \begin{subfigure}[b]{0.8\textwidth}
%     \centering
%     \includegraphics[width=\textwidth]{KGC.png}
%     \caption{Performance comparison of SFT and Base model on different datasets in KGC task.}
%   \end{subfigure}
%   \begin{subfigure}[b]{0.8\textwidth}
%     \centering
%     \includegraphics[width=\textwidth]{QA.png}
%     \caption{Performance comparison of SFT and Base model on different datasets in QA task.}
%   \end{subfigure}
%   % \begin{subfigure}[b]{0.6\textwidth}
%   %   \centering
%   %   \includegraphics[width=\textwidth]{SFT.png}
%   %   \caption{Performance comparison of SFT and Base model on different datasets in T2T task.}
%   % \end{subfigure}
%   % \caption{Performance comparison of SFT and Base model on different datasets in KGC, QA, and T2T tasks.}
%       \label{fig:Base_SFT_Stack_Chart}
% \end{figure*}

% \begin{figure}[h]
%      \centering     \includegraphics[scale=0.145]{SFT.png}
%     \caption{Performance comparison of SFT and Base model on different datasets in T2T task.
%     }
%      \label{fig:SFT}
% \end{figure}

\subsection{RQ4: Comparison Across Model Scales} \label{app:large_models} 
\textbf{Experiment Design.} We conduct a comprehensive comparison of larger LLMs, including Qwen1.5-7B and Qwen3-32B in both their base and SFT variants, together with three strong closed-source LLMs, i.e., GPT-4, Claude-sonnet-4.5, and Gemini-2.5-flash. These models are evaluated on three knowledge-driven tasks across four domain-specific datasets. 

\begin{figure}[t]
     \centering     \includegraphics[width=\linewidth]{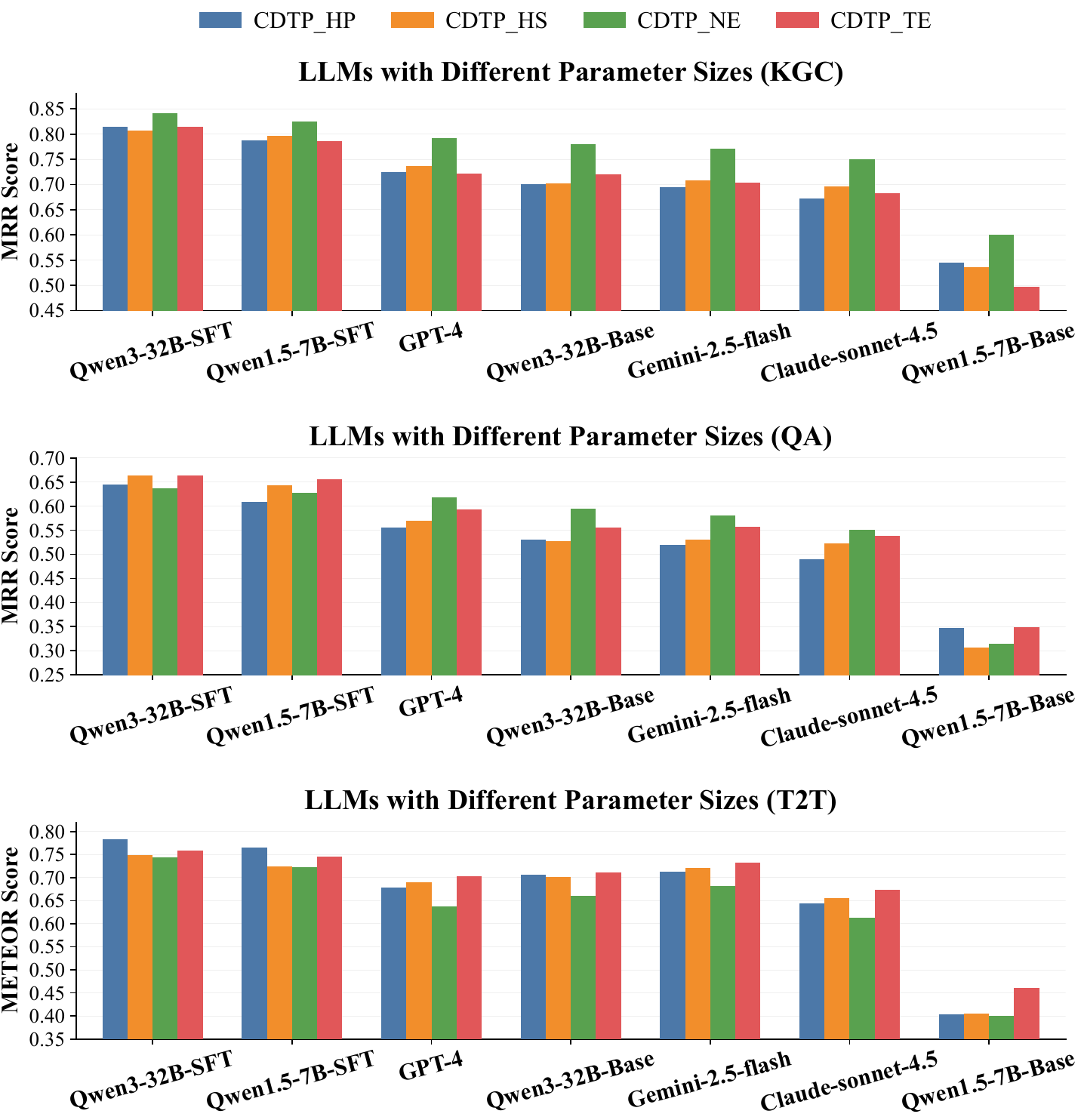}
    \caption{Comparison of LLMs with different parameter sizes.
    }
     \label{fig:LMs_different_parameter_sizes}
     \vspace{-12pt}
\end{figure}

\textbf{Experimental Results.} Figure~\ref{fig:LMs_different_parameter_sizes} presents the comparative results. We have the following observations.

\textbf{\underline{Observation \ding{188}}}: \textbf{Alignment matters more than scale alone.}  
The overall ranking,
Qwen3-32B-SFT $>$ Qwen1.5-7B-SFT $>$ GPT-4 $>$ Qwen3-32B-Base $>$ Gemini-2.5-flash $>$ Claude-sonnet-4.5 $>$ Qwen1.5-7B-Base,
shows that scale alone does not guarantee superior performance. Although Qwen3-32B-Base improves over Qwen1.5-7B-Base, it remains inferior to GPT-4, whereas both SFT variants consistently surpass GPT-4. At the same time, Qwen3-32B-SFT delivers the best overall performance, indicating that the benefits of scale are realized only after effective task- and language-specific alignment.

\textbf{\underline{Observation \ding{189}}}: \textbf{SFT yields larger gains for KGC than for QA.}  
The improvements brought by supervised fine-tuning are systematically larger for KGC than for QA, suggesting that structured knowledge completion benefits more directly from fine-tuning than open-ended question answering.

\subsection{RQ5: Robustness under OOD Data}\label{subsec:robustness}
\textbf{Experiment Design.} We conduct the robustness experiments across three tasks with specialized datasets to verify the robustness of LLMs Yi-9B on OOD data \cite{liu2023good}, before and after SFT using our proposed data. For the KGC task, we select YAGO3-10 \cite{wang2024made} as the OOD dataset and evaluate performance using the MRR metric.
For the QA task, HotpotQA \cite{yang-etal-2018-hotpotqa} is employed, with evaluation based on the F1 Score.
For the T2T task, we adopt the WebNLG dataset \cite{gardent2017webnlg}, using METEOR as the evaluation metric.

\begin{figure}[t]
     \centering     \includegraphics[scale=0.32]{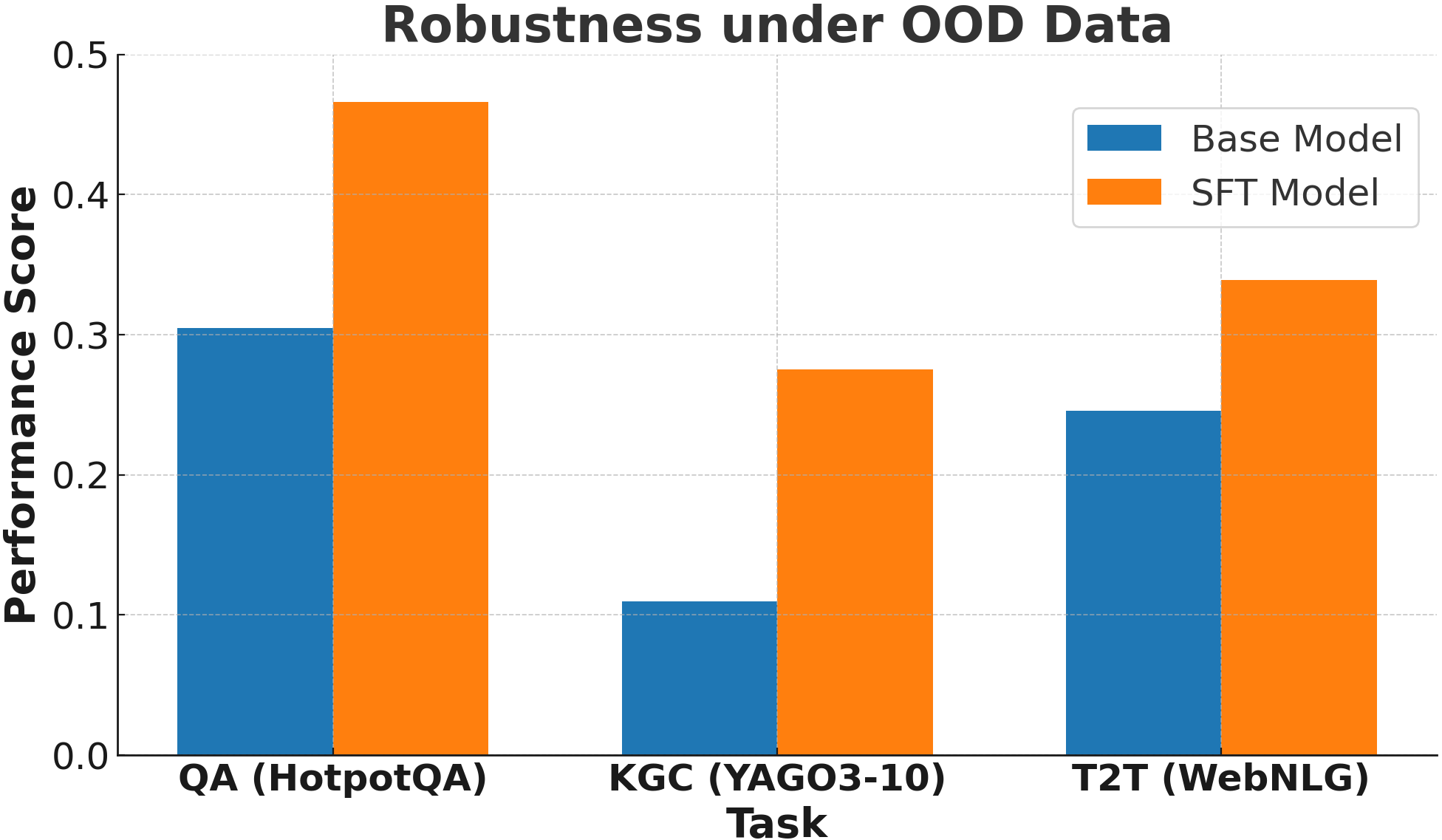}
     \vspace{-2pt}
    \caption{Comparison of robustness between base models and SFT models under the OOD data.
    }
     \label{fig:ood}
     \vspace{-6pt}
\end{figure}

\textbf{Experimental Results.} Figure~\ref{fig:ood} presents the performance comparison across three tasks on three OOD datasets, before and after SFT. We have the following key observations.

\textbf{\underline{Observation~\ding{190}}: Improved OOD performance after fine-tuning.}
As shown in Figure~\ref{fig:ood}, SFT consistently improves performance on OOD data across KGC, QA, and T2T. The gains indicate better generalization to unseen entities, questions, and text patterns, demonstrating both the robustness of LLMs and the effectiveness of the proposed CDTP datasets.

\textbf{\underline{Observation~\ding{191}}: Task-specific OOD robustness.}
Although SFT improves all OOD tasks, the gains vary by task. QA shows the largest improvement, suggesting enhanced reasoning and contextual understanding; KGC benefits substantially through better structured-knowledge generalization; and T2T achieves moderate but stable gains in text generation. These differences highlight the need for task-specific fine-tuning strategies.

\section{Conclusion}

In this paper, we propose a comprehensive benchmark for evaluating LLMs based on the newly constructed Chinese Data-Text Pair (CDTP) dataset. Specifically, CDTP comprises over 7 million meticulously aligned text pairs, each consisting of unstructured text coupled with one or more corresponding triples, alongside a total of 15 million triples spanning four critical domains: History and Politics, Humanities and Society, Technology and Economics, and Nature and Environment. CDTP supports tasks like KGC, T2T, and QA, with a special emphasis on the unique linguistic challenges of Chinese. Furthermore, we conduct rigorous evaluations through extensive experiments and ablation studies to comprehensively assess the effectiveness, SFT, and robustness of the benchmark. This dataset offers a comprehensive benchmark for evaluating reasoning, generalization, and structured knowledge comprehension in LLMs, paving the way for AGI. 

% While CDTP provides a large-scale and strictly aligned Chinese data-to-text benchmark, several limitations remain.

% \begin{itemize}[left=0cm]
%     \item Coverage Bias. 
%     CDTP is mainly constructed from encyclopedic sources and therefore inevitably inherits source-specific biases. Although it covers four broad domains, long-tail entities, emerging facts, and highly specialized knowledge may still be insufficiently represented.

%     \item Generalization Boundary. 
%     The current benchmark validates model performance on in-domain settings and selected OOD scenarios, but its transferability to entirely unseen application scenarios and multimodal settings remains insufficiently examined. Further evaluation is needed to assess whether CDTP-based fine-tuning can robustly generalize beyond text-centered knowledge tasks.

%     \item Task and Linguistic Scope.
%     CDTP focuses on structured factual knowledge through triple-text alignment and evaluates three knowledge-driven tasks: KGC, QA, and T2T. As a result, it is less suited for assessing deeper Chinese language understanding abilities, such as pragmatic inference, metaphorical reasoning, discourse-level coherence, and culturally embedded interpretation.
% \end{itemize}

\bibliography{aaai2027}

@String{Computing = "Computing" }

@article{xu2023superclue,
  title={Superclue: A comprehensive chinese large language model benchmark},
  author={Xu, Liang and Li, Anqi and Zhu, Lei and Xue, Hang and Zhu, Changtai and Zhao, Kangkang and He, Haonan and Zhang, Xuanwei and Kang, Qiyue and Lan, Zhenzhong},
  journal={arXiv preprint arXiv:2307.15020},
  year={2023}
}

@article{huang2024c,
  title={C-eval: A multi-level multi-discipline chinese evaluation suite for foundation models},
  author={Huang, Yuzhen and Bai, Yuzhuo and Zhu, Zhihao and Zhang, Junlei and Zhang, Jinghan and Su, Tangjun and Liu, Junteng and Lv, Chuancheng and Zhang, Yikai and Fu, Yao and others},
  journal={Advances in Neural Information Processing Systems},
  pages={62991--63010},
  year={2023}
}

@article{pan2024unifying,
  title={Unifying large language models and knowledge graphs: A roadmap},
  author={Pan, Shirui and Luo, Linhao and Wang, Yufei and Chen, Chen and Wang, Jiapu and Wu, Xindong},
  journal={IEEE Transactions on Knowledge and Data Engineering},
  volume={36},
  number={7},
  pages={3580--3599},
  year={2024}
}

@article{chang2024survey,
  title={A survey on evaluation of large language models},
  author={Chang, Yupeng and Wang, Xu and Wang, Jindong and Wu, Yuan and Yang, Linyi and Zhu, Kaijie and Chen, Hao and Yi, Xiaoyuan and Wang, Cunxiang and Wang, Yidong and others},
  journal={ACM Transactions on Intelligent Systems and Technology},
  volume={15},
  number={3},
  pages={1--45},
  year={2024}
}

@article{wang2024large,
  title={Large language models-guided dynamic adaptation for temporal knowledge graph reasoning},
  author={Wang, Jiapu and Kai, Sun and Luo, Linhao and Wei, Wei and Hu, Yongli and Liew, Alan Wee-Chung and Pan, Shirui and Yin, Baocai},
  journal={Advances in Neural Information Processing Systems},
  volume={37},
  pages={8384--8410},
  year={2024}
}

@article{wang2024made,
  title={{MADE}: Multicurvature Adaptive Embedding for Temporal Knowledge Graph Completion},
  author={Wang, Jiapu and Wang, Boyue and Gao, Junbin and Pan, Shirui and Liu, Tengfei and Yin, Baocai and Gao, Wen},
  journal={IEEE Transactions on Cybernetics},
  volume={54},
  number={10},
  pages={5818--5831},
  year={2024},
}

@article{pei2019towards,
  title={Towards artificial general intelligence with hybrid Tianjic chip architecture},
  author={Pei, Jing and Deng, Lei and Song, Sen and Zhao, Mingguo and Zhang, Youhui and Wu, Shuang and Wang, Guanrui and Zou, Zhe and Wu, Zhenzhi and He, Wei and others},
  journal={Nature},
  volume={572},
  number={7767},
  pages={106--111},
  year={2019}
}

@article{fei2022towards,
  title={Towards artificial general intelligence via a multimodal foundation model},
  author={Fei, Nanyi and Lu, Zhiwu and Gao, Yizhao and Yang, Guoxing and Huo, Yuqi and Wen, Jingyuan and Lu, Haoyu and Song, Ruihua and Gao, Xin and Xiang, Tao and others},
  journal={Nature Communications},
  volume={13},
  number={1},
  pages={3094},
  year={2022}
}

@article{liu2022chinese,
  title={Chinese named entity recognition: The state of the art},
  author={Liu, Pan and Guo, Yanming and Wang, Fenglei and Li, Guohui},
  journal={Neurocomputing},
  volume={473},
  pages={37--53},
  year={2022}
}

@article{zhu2022improving,
  title={Improving Chinese named entity recognition by large-scale syntactic dependency graph},
  author={Zhu, Peng and Cheng, Dawei and Yang, Fangzhou and Luo, Yifeng and Huang, Dingjiang and Qian, Weining and Zhou, Aoying},
  journal={IEEE/ACM Transactions on Audio, Speech, and Language Processing},
  volume={30},
  pages={979--991},
  year={2022}
}

@article{wang2023survey,
  title={A survey on temporal knowledge graph completion: Taxonomy, progress, and prospects},
  author={Wang, Jiapu and Wang, Boyue and Qiu, Meikang and Pan, Shirui and Xiong, Bo and Liu, Heng and Luo, Linhao and Liu, Tengfei and Hu, Yongli and Yin, Baocai and others},
  journal={arXiv preprint arXiv:2308.02457},
  year={2023}
}

@inproceedings{yang-etal-2018-hotpotqa,
    title = "{H}otpot{QA}: A Dataset for Diverse, Explainable Multi-hop Question Answering",
    author = "Yang, Zhilin  and
      Qi, Peng  and
      Zhang, Saizheng  and
      Bengio, Yoshua  and
      Cohen, William  and
      Salakhutdinov, Ruslan  and
      Manning, Christopher D.",
    booktitle = "Proceedings of the Conference on Empirical Methods in Natural Language Processing",
    year = "2018",
    pages = "2369--2380"
}

@article{glm2024chatglm,
  title={Chatglm: A family of large language models from glm-130b to glm-4 all tools},
  author={GLM, Team and Zeng, Aohan and Xu, Bin and Wang, Bowen and Zhang, Chenhui and Yin, Da and Zhang, Dan and Rojas, Diego and Feng, Guanyu and Zhao, Hanlin and others},
  journal={arXiv preprint arXiv:2406.12793},
  year={2024}
}

@article{young2024yi,
  title={Yi: Open foundation models by 01. ai},
  author={Young, Alex and Chen, Bei and Li, Chao and Huang, Chengen and Zhang, Ge and Zhang, Guanwei and Wang, Guoyin and Li, Heng and Zhu, Jiangcheng and Chen, Jianqun and others},
  journal={arXiv preprint arXiv:2403.04652},
  year={2024}
}

@article{bai2023qwen,
  title={Qwen technical report},
  author={Bai, Jinze and Bai, Shuai and Chu, Yunfei and Cui, Zeyu and Dang, Kai and Deng, Xiaodong and Fan, Yang and Ge, Wenbin and Han, Yu and Huang, Fei and others},
  journal={arXiv preprint arXiv:2309.16609},
  year={2023}
}

@article{yang2023baichuan,
  title={Baichuan 2: Open large-scale language models},
  author={Yang, Aiyuan and Xiao, Bin and Wang, Bingning and Zhang, Borong and Bian, Ce and Yin, Chao and Lv, Chenxu and Pan, Da and Wang, Dian and Yan, Dong and others},
  journal={arXiv preprint arXiv:2309.10305},
  year={2023}
}

@article{bi2024deepseek,
  title={Deepseek llm: Scaling open-source language models with longtermism},
  author={Bi, Xiao and Chen, Deli and Chen, Guanting and Chen, Shanhuang and Dai, Damai and Deng, Chengqi and Ding, Honghui and Dong, Kai and Du, Qiushi and Fu, Zhe and others},
  journal={arXiv preprint arXiv:2401.02954},
  year={2024}
}

@article{javaheripi2023phi,
  title={Phi-2: The surprising power of small language models},
  author={Javaheripi, Mojan and Bubeck, S{\'e}bastien and Abdin, Marah and Aneja, Jyoti and Bubeck, Sebastien and Mendes, Caio C{\'e}sar Teodoro and Chen, Weizhu and Del Giorno, Allie and Eldan, Ronen and Gopi, Sivakanth and others},
  journal={Microsoft Research Blog},
  volume={1},
  number={3},
  pages={3},
  year={2023}
}

@article{achiam2023gpt,
  title={Gpt-4 technical report},
  author={Achiam, Josh and Adler, Steven and Agarwal, Sandhini and Ahmad, Lama and Akkaya, Ilge and Aleman, Florencia Leoni and Almeida, Diogo and Altenschmidt, Janko and Altman, Sam and Anadkat, Shyamal and others},
  journal={arXiv preprint arXiv:2303.08774},
  year={2023}
}

@inproceedings{dong-etal-2024-abilities,
    title = "How Abilities in Large Language Models are Affected by Supervised Fine-tuning Data Composition",
    author = "Dong, Guanting  and
      Yuan, Hongyi  and
      Lu, Keming  and
      Li, Chengpeng  and
      Xue, Mingfeng  and
      Liu, Dayiheng  and
      Wang, Wei  and
      Yuan, Zheng  and
      Zhou, Chang  and
      Zhou, Jingren",
    booktitle = "Proceedings of the 62nd Annual Meeting of the Association for Computational Linguistics",
    year = "2024",
    pages = "177--198"
}

@inproceedings{instag2025,
  title={\#{INSTAG}: Instruction tagging for analyzing supervised fine-tuning of large language models},
  author={Lu, Keming and Yuan, Hongyi and Yuan, Zheng and Lin, Runji and Lin, Junyang and Tan, Chuanqi and Zhou, Chang and Zhou, Jingren},
  booktitle = {International Conference on Learning Representations},
  pages = {1--10},
  year      = {2024}
}

@inproceedings{liu2023good,
  title={Good-d: On unsupervised graph out-of-distribution detection},
  author={Liu, Yixin and Ding, Kaize and Liu, Huan and Pan, Shirui},
  booktitle={Proceedings of the  ACM International Conference on Web Search and Data Mining},
  pages={339--347},
  year={2023}
}

@inproceedings{gardent2017webnlg,
  title={{The WebNLG Challenge}: Generating Text from RDF Data},
  author={Gardent, Claire and Shimorina, Anastasia and Narayan, Shashi and Perez-Beltrachini, Laura},
  booktitle={Proceedings of the International Conference on Natural Language Generation},
  pages={124--133},
  year={2017},
}

@inproceedings{rasley2020deepspeed,
  title={Deepspeed: System optimizations enable training deep learning models with over 100 billion parameters},
  author={Rasley, Jeff and Rajbhandari, Samyam and Ruwase, Olatunji and He, Yuxiong},
  booktitle={Proceedings of the 26th ACM SIGKDD international conference on knowledge discovery \& data mining},
  pages={3505--3506},
  year={2020}
}

@inproceedings{li2021triple,
  title={Triple-to-text generation with an anchor-to-prototype framework},
  author={Li, Ziran and Lin, Zibo and Ding, Ning and Zheng, Hai-Tao and Shen, Ying},
  booktitle={Proceedings of the Twenty-Ninth International Conference on International Joint Conferences on Artificial Intelligence},
  pages={3780--3786},
  year={2021}
}

@article{he2024g,
  title={G-retriever: Retrieval-augmented generation for textual graph understanding and question answering},
  author={He, Xiaoxin and Tian, Yijun and Sun, Yifei and Chawla, Nitesh and Laurent, Thomas and LeCun, Yann and Bresson, Xavier and Hooi, Bryan},
  journal={Advances in Neural Information Processing Systems},
  volume={37},
  pages={132876--132907},
  year={2024}
}

@article{yan2025prompting,
  title={Prompting Large Language Models with Partial Knowledge for Answering Questions with Unseen Entities},
  author={Yan, Zhichao and Wang, Jiapu and Chen, Jiaoyan and Wang, Yanyan and Tan, Hongye and Liang, Jiye and Li, Xiaoli and Li, Ru and Pan, Jeff Z},
  journal={arXiv preprint arXiv:2508.01290},
  year={2025}
}

@article{li2025matching,
  title={From matching to generation: A survey on generative information retrieval},
  author={Li, Xiaoxi and Jin, Jiajie and Zhou, Yujia and Zhang, Yuyao and Zhang, Peitian and Zhu, Yutao and Dou, Zhicheng},
  journal={ACM Transactions on Information Systems},
  volume={43},
  number={3},
  pages={1--62},
  year={2025}
}

@article{li2024pre,
  title={Pre-trained language models for text generation: A survey},
  author={Li, Junyi and Tang, Tianyi and Zhao, Wayne Xin and Nie, Jian-Yun and Wen, Ji-Rong},
  journal={ACM Computing Surveys},
  volume={56},
  number={9},
  pages={1--39},
  year={2024}
}

@inproceedings{wang2024ime,
  title={{IME}: Integrating multi-curvature shared and specific embedding for temporal knowledge graph completion},
  author={Wang, Jiapu and Cui, Zheng and Wang, Boyue and Pan, Shirui and Gao, Junbin and Yin, Baocai and Gao, Wen},
  booktitle={Proceedings of the ACM Web Conference 2024},
  pages={1954--1962},
  year={2024}
}

@inproceedings{li2023cmmlu,
  title={{CMMLU}: Measuring massive multitask language understanding in chinese},
  author={Li, Haonan and Zhang, Yixuan and Koto, Fajri and Yang, Yifei and Zhao, Hai and Gong, Yeyun and Duan, Nan and Baldwin, Timothy},
  booktitle={Findings of the Association for Computational Linguistics},
  pages={11260--11285},
  year={2024}
}

@ARTICLE{yan2024atomic,
author={Yan, Zhichao and Wang, Jiapu and Chen, Jiaoyan and Li, Xiaoli and Liang, Jiye and Li, Ru and Pan, Jeff Z.},
journal={ IEEE Transactions on Knowledge and Data Engineering },
title={Atomic Fact Decomposition Helps Attributed Question Answering },
year={2025},
volume={37},
number={12},
pages={6959-6972},
}

@article{cai2024internlm2,
  title={Internlm2 technical report},
  author={Cai, Zheng and Cao, Maosong and Chen, Haojiong and Chen, Kai and Chen, Keyu and Chen, Xin and Chen, Xun and Chen, Zehui and Chen, Zhi and Chu, Pei and others},
  journal={arXiv preprint arXiv:2403.17297},
  year={2024}
}

@article{grattafiori2024llama,
  title={The llama 3 herd of models},
  author={Grattafiori, Aaron and Dubey, Abhimanyu and Jauhri, Abhinav and Pandey, Abhinav and Kadian, Abhishek and Al-Dahle, Ahmad and Letman, Aiesha and Mathur, Akhil and Schelten, Alan and Vaughan, Alex and others},
  journal={arXiv preprint arXiv:2407.21783},
  year={2024}
}

@article{wang2023multi,
  title={Multi-concept representation learning for knowledge graph completion},
  author={Wang, Jiapu and Wang, Boyue and Gao, Junbin and Hu, Yongli and Yin, Baocai},
  journal={ACM Transactions on Knowledge Discovery from Data},
  volume={17},
  number={1},
  pages={1--19},
  year={2023}
}

@inproceedings{sun2025large,
  title={Large-Small Model Synergy with Multimodal Fine-Grained Heuristics for Knowledge-Based Visual Question Answering},
  author={Sun, Zhongfan and Guo, Kan and Hu, Yongli and Tian, Daxin and Gao, Qingqing and Wang, Jiapu and Gao, Junbin and Sun, Yanfeng and Yin, Baocai},
  booktitle={Proceedings of the 33rd ACM International Conference on Multimedia},
  pages={935--944},
  year={2025}
}

@article{sokolova2009systematic,
  title={A systematic analysis of performance measures for classification tasks},
  author={Sokolova, Marina and Lapalme, Guy},
  journal={Information processing \& management},
  volume={45},
  number={4},
  pages={427--437},
  year={2009}
}

@inproceedings{BLEU,
  author       = {Kishore Papineni and
                  Salim Roukos and
                  Todd Ward and
                  Wei{-}Jing Zhu},
  title        = {Bleu: a Method for Automatic Evaluation of Machine Translation},
  booktitle    = {Proceedings of the 40th Annual Meeting of the Association for Computational Linguistics},
  pages        = {311--318},
  year         = {2002}
}

@inproceedings{ROUGE,
    title = "{ROUGE}: A Package for Automatic Evaluation of Summaries",
    author = "Lin, Chin-Yew",
    booktitle = "Text Summarization Branches Out",
    month = jul,
    year = "2004",
    pages = "74--81"
}

@inproceedings{METEOR,
    title = "{METEOR}: An Automatic Metric for {MT} Evaluation with Improved Correlation with Human Judgments",
    author = "Banerjee, Satanjeev  and
      Lavie, Alon",
    booktitle = "Proceedings of the {ACL} Workshop on Intrinsic and Extrinsic Evaluation Measures for Machine Translation and/or Summarization",
    year = "2005",
    pages = "65--72"
}

@inproceedings{
zhuo2026knowledge,
title={Knowledge Reasoning Language Model: Unifying Knowledge and Language for Inductive Knowledge Graph Reasoning},
author={Xingrui Zhuo and Jiapu Wang and Gongqing Wu and Zhongyuan Wang and Jichen Zhang and Shirui Pan and Xindong Wu},
booktitle={The Fourteenth International Conference on Learning Representations},
year={2026}
}

@inproceedings{DBLP:conf/aaai/HuangBHZW26,
  author       = {Manzong Huang and
                  Chenyang Bu and
                  Yi He and
                  Xingrui Zhuo and
                  Xindong Wu},
  title        = {Relink: Constructing Query-Driven Evidence Graph On-the-Fly for GraphRAG},
  booktitle    = {Fortieth {AAAI} Conference on Artificial Intelligence},
  pages        = {31202--31210},
  year         = {2026}
}

@ARTICLE{Jing2026knowledge,
  author={Jing, Xiaonan and Wu, Gongqing and Zhuo, Xingrui},
  journal={IEEE Transactions on Audio, Speech and Language Processing}, 
  title={Generation-Then-Discrimination: Hierarchical Entity-Relation Decoder for Relational Triplet Extraction}, 
  year={2026},
  volume={34},
  number={1},
  pages={1826-1839}
}

%\clearpage
%\large \section{Appedix}

\end{document}